\pdfoutput=1

\documentclass[%
aip,
amsmath,amssymb,
reprint,%
]{revtex4-1}

\usepackage{graphicx}
\usepackage{dcolumn}
\usepackage{bm}

\usepackage[utf8]{inputenc}
\usepackage[T1]{fontenc}
\usepackage{mathptmx}
\usepackage{etoolbox}

\usepackage{xcolor}
\usepackage{soul}

\usepackage{hyperref}
\hypersetup{
	colorlinks=true,
	linkcolor=blue,
	citecolor=blue,
	urlcolor=blue
}

\makeatletter
\def\@email#1#2{%
	\endgroup
	\patchcmd{\titleblock@produce}
	{\frontmatter@RRAPformat}
	{\frontmatter@RRAPformat{\produce@RRAP{*#1\href{mailto:#2}{#2}}}\frontmatter@RRAPformat}
	{}{}
}%
\makeatother
\begin{document}
	
	\preprint{AIP/123-QED}
	\title[Adaptive Reservoir Computing]{Evolutionary Optimization Reveals Structural Constraints on Reservoir Architecture for Spatiotemporal Chaos }

	\author{Nima Dehghani}
	\email{nima.dehghani@mit.edu}
	\altaffiliation
	[Current Address: ]{McGovern Institute for Brain Research, MIT
		\\
		Context \& Overview: \\
		\url{https://neurovium.science/posts/pblog-Adaptive-reservoir} \\ 
		Code \& Experiments: \\
		\url{https://github.com/neurovium/AdaptiveReservoirComputing}
	}
	\affiliation{ 
		Department of Physics, MIT
	}
	\affiliation{ 
		McGovern Institute for Brain Research, MIT 
	}

	\date{\today}

	\begin{abstract}
		Biological systems maintain function in fluctuating environments by transforming past stimulation into internal dynamical states that support future-oriented responses. Reservoir computing provides a computational analogue, but the standard framework usually treats the recurrent substrate as a fixed random network and trains only the readout. Here we ask how the recurrent substrate itself changes when reservoir architecture is placed under evolutionary selection for prediction. Using the Kuramoto--Sivashinsky equation as a testbed for spatiotemporal chaos, we evolved reservoirs over five construction hyperparameters: size, connectivity degree, spectral radius, input scaling, and readout regularization. Evolutionary optimization reduced prediction error at the population level, extended the low-error forecast horizon, and organized the reservoir design space along a diminishing-return size--efficiency frontier. Structural analyses revealed that evolved reservoirs remained within a conserved stochastic-block-model-like spectral envelope while showing directional refinement of low-eigenvalue modes, locking of macroscopic modularity to a narrow intermediate band, and exponential pruning of connection cost within that band. Pareto analysis showed that elite predictive reservoirs occupied a horizontal floor in the cost--modularity plane, indicating that accuracy and structural efficiency were achieved jointly rather than through a simple trade-off. Together, these findings show that evolutionary optimization does not merely reduce prediction error, but exposes interpretable structural constraints on the recurrent substrate itself, stabilizing a task-suitable dynamical class and refining the architectural degrees of freedom most relevant for prediction. Evolutionary reservoir computing therefore provides a bio-inspired framework for studying how predictive demands shape adaptive dynamical networks across machine learning, complex systems, and biological computation.
	\end{abstract}


\maketitle

\section{\label{sec:intro}Introduction}

Many biological systems persist in fluctuating environments by anticipating, rather than merely reacting to, temporal structure \cite{tagkopoulos2008predictivebehaviorwithin,freddolino2012beyondhomeostasisa,deans2021biologicalpresciencethe}. Cells, neural circuits, and regulatory networks are continuously driven by external signals whose future states are only partially predictable \cite{freddolino2012beyondhomeostasisa,sussillo2014neuralcircuits}. In this sense, adaptation is not only a matter of response, but also of prediction. Maintaining function under such conditions requires internal dynamics that transform past stimulation into informative present states. This predictive capacity need not take the form of an explicit symbolic or formal model of the external world. Instead, molecular reactions, recurrent connectivity, feedback loops, and internal state variables can provide dynamical memory through which past inputs condition future responses. In this view, prediction emerges from the organization of a physical or biological substrate, rather than from a separately represented model \cite{freddolino2012beyondhomeostasisa,gabaldasagarra2018recurrencebasedinformationprocessing,seoane2019evolutionaryaspectsof,dehghani2024physicalcomputing,dehghani2018multiscalecontrol,dehghani2024cellularcomplexity}.

Reservoir computing provides a useful abstraction for studying this principle. A recurrent dynamical system maps an input time series into a high-dimensional state space, while a trained readout extracts the desired output \cite{jaeger2001echo,JaegerHaas2004esn,maass2002realtime,lukosevicius2009reservoircomputing,maass2007liquidcomputing,zhang2023asurveyon}. Because the recurrent reservoir is often fixed after initialization, computation depends on how the intrinsic dynamics of the substrate transform input history into a useful state representation. This makes reservoir computing especially relevant for chaotic timeseries prediction and control, where small errors grow rapidly and successful forecasting requires the model to preserve the temporal and geometrical structure of the underlying dynamics over a finite horizon \cite{pathak2017usingmachinelearning,pathak2018modelfreepredictionof,pathak2018hybridforecastingof,Krishnagopal2020chaosreservoir,Kent2024chaoscontrol,Griffith2019forecastchaos}. The Kuramoto--Sivashinsky equation \cite{Kuramoto1978KS,Sivashinsky1977KS}, a canonical model of spatiotemporal chaos combining nonlinear advection, instability, and higher-order dissipation, provides a stringent testbed for this question \cite{rost1995aparticlemodel,tomlin2018nonlineardynamicsof,pathak2018modelfreepredictionof}.

\begin{figure*}
    \centering
    \includegraphics[width=\linewidth]{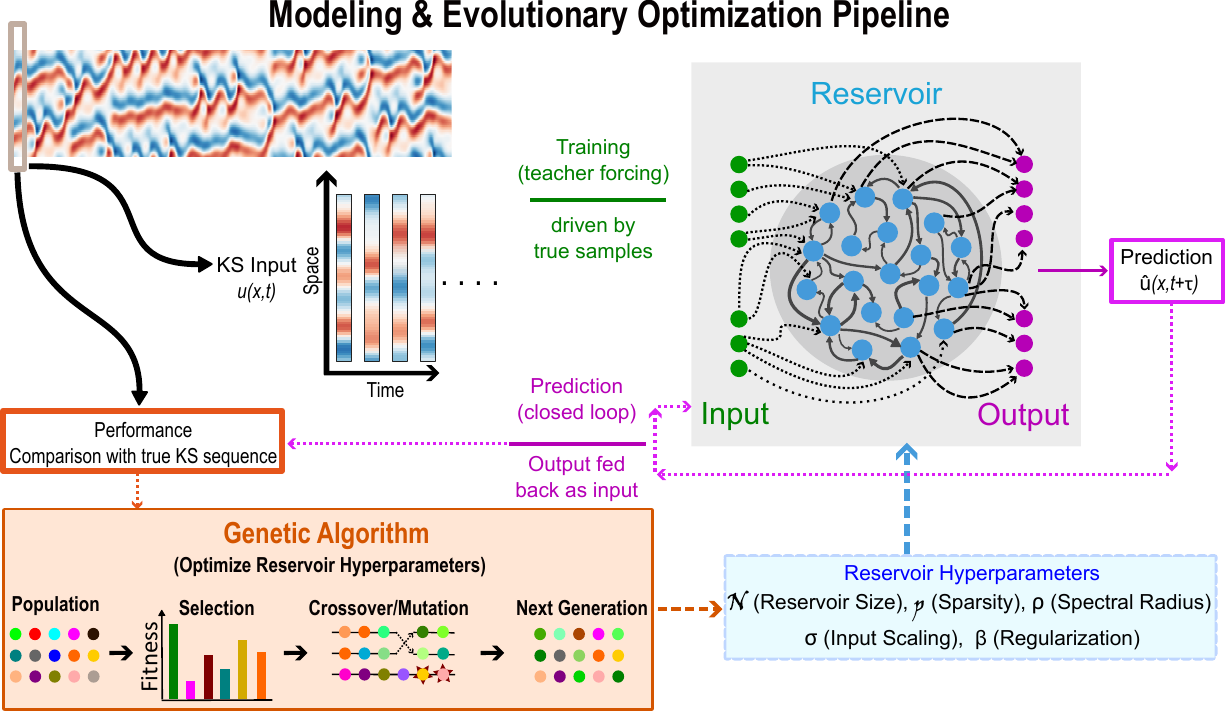}
        \caption{
        \textbf{End‑to‑End Reservoir Optimization Loop.}
        Kuramoto‑Sivashinsky timeseries condition the \textit{reservoir} during supervised teacher training so that the internal state encodes the flow dynamics. During evaluation, the model is decoupled from the dataset and performs closed‑loop prediction, generating its own next‑step inputs; the resulting forecast error over many steps quantifies stability and accuracy. Those errors drive a genetic algorithm that explores the hyperparameter space, promoting configurations that yield robust, long‑range predictions.
        }
    \label{fig:overview}
\end{figure*}

Previous work has shown that reservoir performance depends not only on the trained readout, but also on reservoir structure, operating regime, topology, size, spectral radius, excitatory--inhibitory balance, and other hyperparameters \cite{carroll2019networkstructure,carroll2020edgeofchaos,haluszczynski2019goodandbad,kitayama2022guidingprincipleof,Srinivasan2025}. Reservoirs have been used to forecast spatially extended dynamical systems, including spatiotemporal chaos and excitable media \cite{pathak2018modelfreepredictionof,zimmermann2018observing,Krishnagopal2020chaosreservoir}, and evolutionary or metaheuristic methods have been used to tune reservoir hyperparameters and architectural features \cite{ferreira2009geneticalgorithmfor,roeschies2010structureoptimizationof,bala2018applicationsofmetaheuristics} (For reviews of optimization using evolutionary approaches of neural networks and reservoir computing, see \cite{Stanley2019neuroevo,Schmidhuber2007evo,basterrech2023revisitingreservoircomputing}). However, these results leave open a more structural question: when predictive performance is selected, what kind of recurrent substrate is favored? In particular, does selection merely tune scalar hyperparameters that improve error, or does it reveal reproducible constraints on the size, spectrum, modularity, and cost of the recurrent network? 

The motivation is therefore both computational and biological: computationally, to determine whether evolutionary optimization can improve chaotic prediction while revealing how capacity, connectivity, and cost are jointly constrained; biologically, to use reservoir computing as a controlled abstraction for asking how adaptive predictive function may emerge from the tuning of dynamical substrates. Here we address this question by evolving reservoir computers for prediction of Kuramoto--Sivashinsky spatiotemporal chaos. A genetic algorithm optimized reservoir size, spectral radius, input scaling, regularization, and connectivity degree, while each candidate reservoir was evaluated by autonomous closed-loop prediction (Fig.~\ref{fig:overview}). We then analyzed not only prediction error, but also forecast horizon, size--efficiency tradeoffs, Laplacian spectra, modularity, and connection cost. This allows us to ask how predictive improvement is expressed in the recurrent substrate itself.

We find that evolutionary optimization improves prediction at the population level, extending the low-error forecast horizon and producing an empirical size--efficiency frontier with diminishing returns. The structural analyses reveal a more specific organization: evolved reservoirs remain within a conserved SBM-like spectral envelope, while selection acts directionally on the lowest part of the Laplacian spectrum; modularity becomes constrained to a narrow intermediate band; and connection cost is pruned within that band. These results suggest that adaptive recurrent prediction is not achieved by unconstrained architectural growth or arbitrary random search. Instead, selection stabilizes a task-suitable dynamical class and refines the degrees of freedom within that class that most strongly affect predictive function.

\section{\label{sec:results}Results}

\subsection{Evolutionary optimization improves reservoir prediction of Kuramoto--Sivashinsky dynamics}

\begin{figure*}
    \centering
    \includegraphics[width=\linewidth]{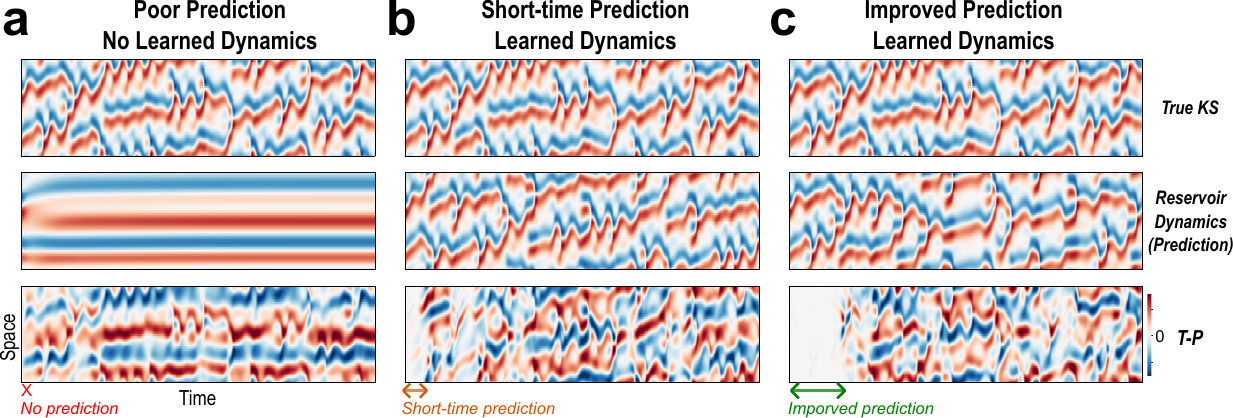}
        \caption{%
        \textbf{Comparison of ground truth, autonomous prediction, and error for three reservoir configurations (a–c).}  
        Top row (T) shows the reference Kuramoto–Sivashinsky trajectory; middle row (P) shows the reservoir run autonomously in a closed-loop fashion (each predicted timestep fed back as the next input); bottom row (T$-$P) shows the pointwise difference between truth and prediction, highlighting spatial–temporal error growth. Column \textbf{(a)} is a poorly configured reservoir with essentially zero useful prediction horizon: its autonomous output does not resemble KS dynamics and errors grow immediately and globally. Column \textbf{(b)} is a reservoir that has learned the short‑time KS dynamics: its autonomous rollouts visually match the true patterns for several timesteps and the error remains small over a short horizon. Column \textbf{(c)} is a well‑tuned reservoir with a substantially longer prediction horizon: it sustains KS‑like dynamics for many steps and shows much slower error accumulation. The difference panels emphasize where and how forecast fidelity breaks down as a function of reservoir quality.}
    \label{fig:ks_prediction_examples}
\end{figure*}

We first asked whether evolutionary optimization improves the ability of reservoir computing networks to predict the spatiotemporal dynamics of the Kuramoto--Sivashinsky system. Reservoirs were evolved over five hyperparameters: size, spectral radius, input scaling, regularization, and connectivity degree (see Methods). Each reservoir was evaluated using the composite prediction-error score \(J\), defined in Eq.~\ref{eq:fitness_function}, which combines the normalized mean absolute prediction error with the number of output dimensions whose NRMSE falls below threshold; lower values of \(J\) therefore indicate higher predictive fitness. For an overview of the evolutionary optimization of reservoir and prediction pipeline, see Fig.~\ref{fig:overview} and Fig.~\ref{fig:ks_prediction_examples}.


\subsection{Population-level reduction of composite prediction error}

\begin{figure}
    \centering
    \includegraphics[width=\linewidth]{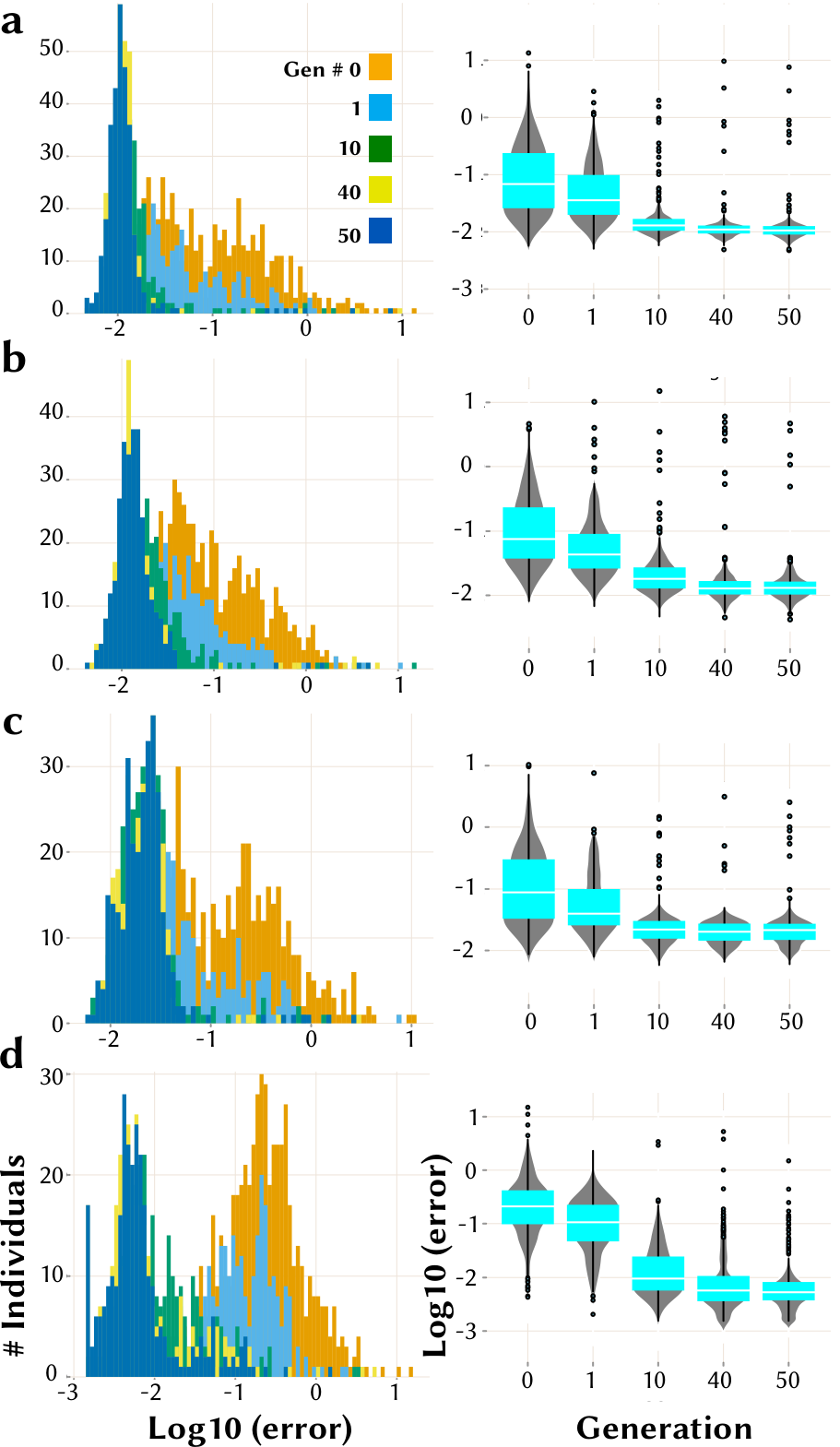}
        \caption{%
        \textbf{Population‑level reduction of composite prediction error \(J\) across representative genetic‑algorithm runs.} 
        Four independent runs (\texttt{RUNS $\#$ 1,3,5,7}) are shown as representative examples in which evolutionary optimization reduces prediction error. For each run the left subpanel displays the histogram/density of the composite fitness score \(J\) across the full population of 300 individual reservoirs at generations \(0,1,10,40,\) and \(50\); the right subpanel summarizes the same progression using distribution‑summary plots of \(\log_{10} J\). The composite score \(J\), defined in Eq.~\ref{eq:fitness_function}, is the ratio of the min–max normalized mean absolute error (NMAE, computed after min–max normalization of predicted and target outputs) to the number of output dimensions whose \(\mathrm{NRMSE}_k\) falls below the threshold \(\varepsilon\); lower \(J\) therefore indicates better predictive fitness. Values are shown on a \(\log_{10}\) scale to make generational shifts in the error distribution visually comparable across runs. In all runs, the distribution of \(J\) shifts rapidly toward lower values, with most improvement occurring early in evolution. Crucially, this improvement is population‑wide: the entire distribution moves toward lower error, the prevalence of high‑error reservoirs declines, and the fraction of low‑error individuals increases. The right‑hand summary panels emphasize the downward shift in central tendency and the reduced spread of later generations. Similar findings across all independent runs demonstrate that evolutionary search reshapes the full population of reservoirs per generation rather than merely discovering isolated elites.
        }%

    \label{fig:j_histograms}
\end{figure}

Across generations, evolutionary optimization produced a systematic reduction in the composite error \(J\) (Fig.~\ref{fig:j_histograms}). Each generation contained 300 reservoirs, allowing us to examine the full distribution of predictive fitness rather than only the best-performing individuals.

In the initial generations, the distributions of \(J\) were broad, reflecting the variability of randomly initialized reservoirs and centered near \(\log_{10} J \approx -1.0\). As evolution progressed, the distributions shifted toward lower values, with late-stage populations converging near \(\log_{10} J \approx -2.0\)---an order-of-magnitude reduction in composite error. Much of this shift occurred in the earliest generations, consistent with a rapid transition from broad exploration of reservoir-parameter space to local refinement around favorable regions: early selection removed poorly performing reservoirs, while later selection refined architectures already occupying low-error regimes.

Importantly, the improvement was not confined to a small elite subset. The full population distribution shifted toward lower composite error, with fewer high-error reservoirs in later generations and a larger fraction of individuals occupying low-error regimes. The spread of \(J\) also narrowed across generations. Evolutionary optimization thus improved both peak prediction and population-level robustness, reshaping the distribution of reservoir performance rather than merely discovering isolated high-performing networks. This pattern was observed in all independent simulation runs.

\subsection{Evolution extends the duration of low-error prediction}

The composite score \(J\) provides a scalar measure of predictive fitness but does not show how prediction error evolves over the forecast horizon. We therefore examined the time-resolved normalized root mean square error (NRMSE) for all reservoirs in a representative run at the beginning and end of evolution. For each reservoir, we computed a threshold-based prediction length, defined as the number of time samples for which \(\mathrm{NRMSE}<0.05\). This threshold was used heuristically to order reservoirs by the duration over which they maintained low prediction error.

\begin{figure*}
    \centering
    \includegraphics[width=\linewidth]{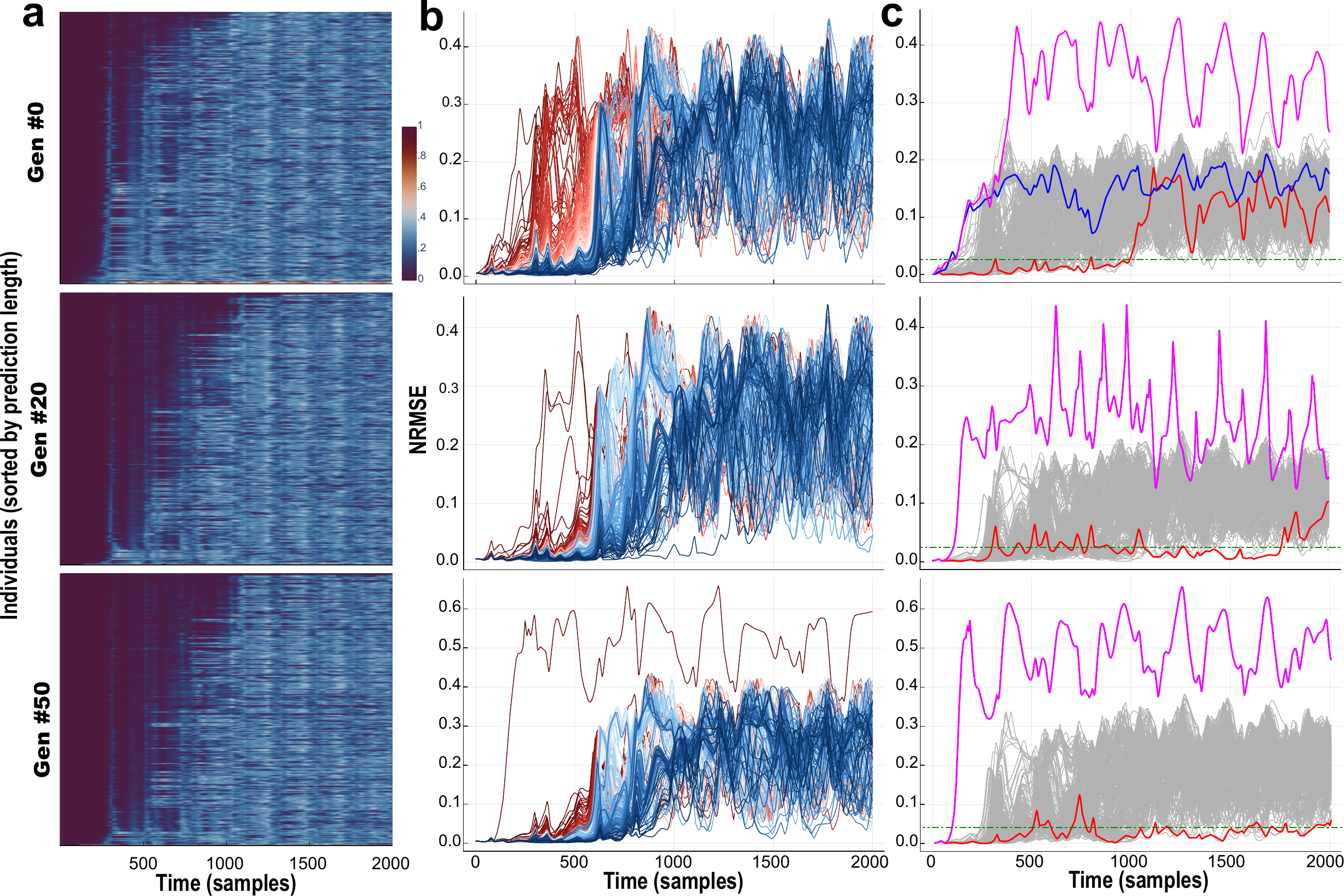}
        \caption{
        \textbf{Evolution extends the duration of low‑error prediction across the reservoir population.} 
        Column \textbf{(a)} shows heatmaps of the time‑resolved normalized root mean square error \(\mathrm{NRMSE}\) for all reservoirs in \texttt{RUN \#4} at Generations 0, 20, and 50. Each row is one reservoir; rows are sorted within each generation by a threshold‑based prediction length, defined as the number of time samples for which \(\mathrm{NRMSE}<0.05\); this ordering places reservoirs with the longest low‑error intervals at the top. In the initial population (Generation 0) many reservoirs exhibit only short low‑error intervals, whereas by Generation 50 a substantially larger fraction maintain low \(\mathrm{NRMSE}\) for extended portions of the forecast horizon. In column \textbf{(b)} individual networks from \texttt{RUN \#2}  Generations 0, 20, and 50 are color‑coded \(\mathrm{NRMSE}\) trajectories with warmer colors indicating poorer performance and cooler colors indicating better performance. These trajectory views make clear that evolution increases the number of reservoirs that retain low prediction error over longer time spans. Column \textbf{(c)}: \texttt{RUN \#6}  Generations 0, 20, and 50, gray lines denote the full population, the red trace marks the reservoir with the longest threshold‑based good prediction length, the blue trace marks the reservoir with the shortest threshold‑based prediction length, the magenta trace marks the reservoir with the largest cumulative NRMSE, and the green dashed line indicates the threshold \(\mathrm{NRMSE}=0.05\). Similar findings across different runs demonstrate that evolutionary optimization not only reduces aggregate fitness \(J\) but also extends the low‑error forecasting interval across the evolved population.}%
    \label{fig:nrmse_temporal}
\end{figure*}

The contrast between initial and evolved populations is clear in the NRMSE heatmaps (Fig.~\ref{fig:nrmse_temporal}.a). Within the Generation 0 population, many reservoirs exhibited only short intervals of low NRMSE, with prediction errors growing relatively early in the forecast horizon. Within the Generation 50 population, a substantially larger fraction of reservoirs maintained low NRMSE for longer intervals. The ordered NRMSE trajectory panels  (Fig.~\ref{fig:nrmse_temporal}.b,Fig.~\ref{fig:nrmse_temporal}.c) show the same pattern at the level of individual error time courses: after evolution, more trajectories remain near the low-error regime for longer portions of the prediction window. Representative Generation 50 trajectories illustrate the range of outcomes within the evolved population, including the reservoir with the longest threshold-based prediction length and reservoirs with substantially larger accumulated error.

Consistent results were observed across independent simulation runs. Together, these analyses show that the reduction in \(J\) corresponds to a time-resolved extension of low-error forecasting across the reservoir population.
 This convergence indicates that the GA successfully navigated the topological landscape to identify an architectural template specifically tuned to the KS system's Lyapunov exponents.

\subsection{Evolution organizes the reservoir empirical size--efficiency frontier}
\label{sec:size_efficiency} 
A central question was how improved prediction related to reservoir size. In principle, better performance could arise simply by increasing the number of reservoir units, thereby expanding the dimensionality and expressive capacity of the recurrent system. Conversely, evolution could favor smaller reservoirs if the dominant effect of optimization were to remove unnecessary structural degrees of freedom. A third possibility is that evolution organizes the population along a trade-off between size and error rather than collapsing it onto a single optimal size. To distinguish among these alternatives, we represented each evaluated reservoir as a point in the size--error plane, with reservoir size \(n_r\) on one axis and log composite prediction error \(\log J\) on the other.

\begin{figure*}
    \centering
    \includegraphics[width=\linewidth]{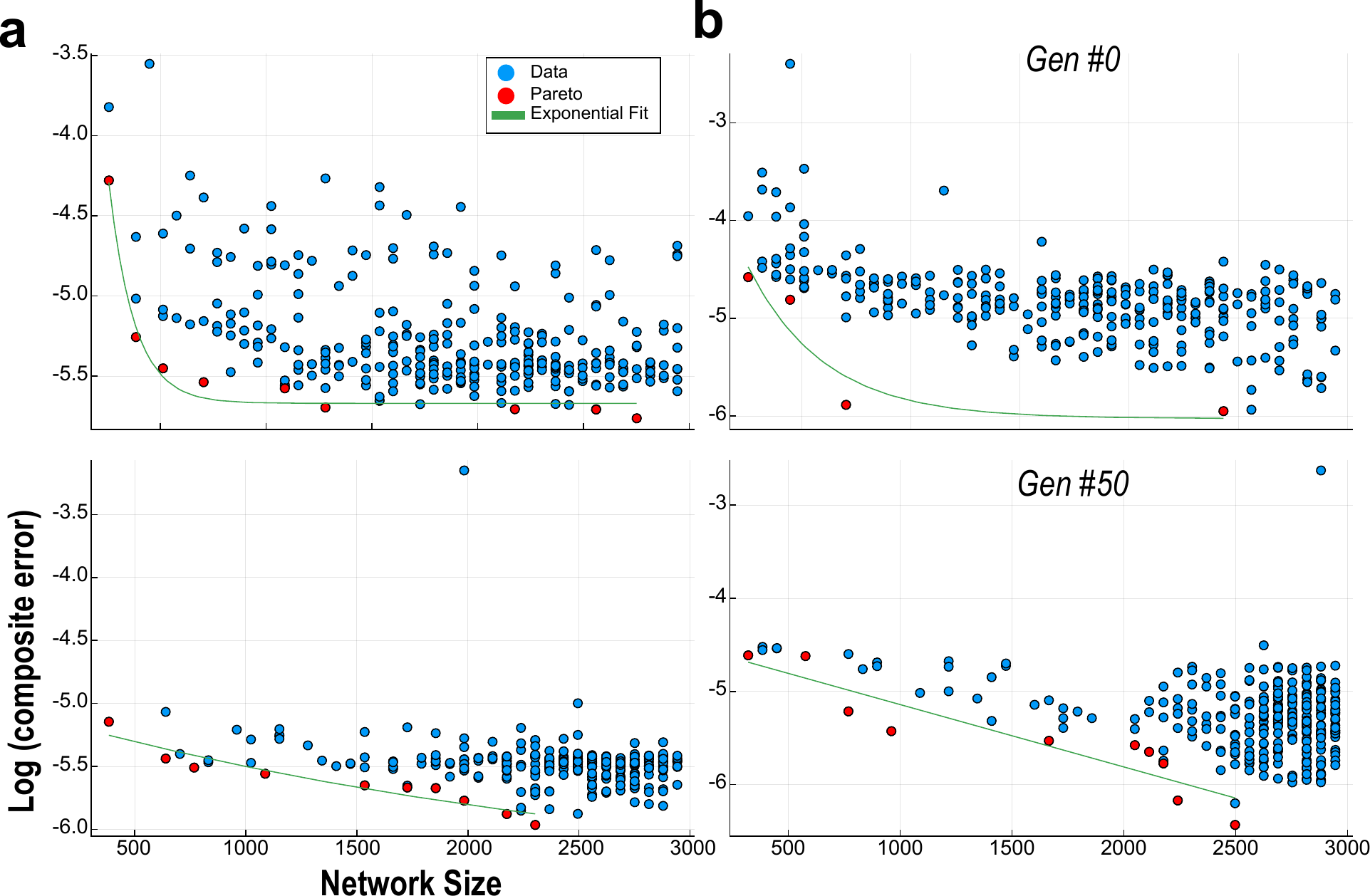}
    \caption{
        \textbf{Evolution of the reservoir empirical size--error frontier across generations.}
        Panels show generation 0 (top row) and generation 50 (bottom row) for \texttt{RUN \#2} (left column) and \texttt{RUN \#4} (right column). 
        Each point corresponds to one evaluated reservoir, with reservoir size \(n_r\) on the horizontal axis and log composite error \(\log J\) on the vertical axis. 
        Red points denote nondominated reservoirs for which no other configuration achieved both smaller size and lower error. 
        The fitted curve provides a descriptive summary of the empirical size-error frontier. 
        At generation 0, the size--error landscape is diffuse, with both small and large reservoirs spanning a broad range of errors. 
        By generation 50, the nondominated points form a more coherent frontier, indicating that evolutionary optimization organizes the reservoir population around a reproducible size--accuracy trade-off. 
        The late-generation frontiers show a diminishing-return structure: larger reservoirs can achieve lower attainable error, but the marginal improvement decreases at larger \(n_r\). 
        Similar frontier shapes across independent runs indicate that this trade-off is a robust feature of the evolved reservoir design space for the Kuramoto--Sivashinsky prediction task.
    }
    \label{fig:size-error}
\end{figure*}

At the initial generation, the size--error landscape was broad and weakly organized (Fig.~\ref{fig:size-error}.Top). Reservoirs of many different sizes produced poor predictions, and large reservoirs were not automatically associated with low error, indicating that size alone was not sufficient to determine predictive performance.

As evolution progressed, the initially diffuse population condensed around a well-defined empirical Pareto frontier. This empirical frontier is identified post hoc from the evolved reservoir population using a dominance criterion (Methods); it is descriptive of the evolved data and is not the output of a separate optimization run. We refer to it throughout as the \emph{empirical size--error frontier} to distinguish it from the multi-objective Pareto analysis presented in Sec.~\ref{sec:multiobj}. By the final generation, the nondominated reservoirs formed a coherent empirical boundary in the size--error plane. This frontier did not indicate that evolution simply minimized reservoir size. Rather, it revealed a structured trade-off: smaller reservoirs could remain Pareto efficient when they provided compact but less accurate solutions, whereas larger reservoirs occupied the low-error region of the frontier when their additional degrees of freedom were paired with favorable dynamical parameters (Fig.~\ref{fig:size-error}.Bottom).

The late-generation frontier showed a diminishing-return pattern. Along the nondominated boundary, increasing reservoir size lowered attainable error, but with diminishing marginal gains at larger \(n_r\). Evolution therefore did not favor larger reservoirs indiscriminately; many large networks remained dominated when their hyperparameters were poorly tuned, while smaller reservoirs persisted on the frontier when they offered compact but less accurate solutions. Efficient performance required the joint tuning of reservoir size, spectral radius, connectivity degree, input scaling, and regularization, rather than the adjustment of size alone.

This pattern was qualitatively reproducible across independent evolutionary runs. Although individual reservoirs and evolutionary trajectories differed across runs, the final empirical size--error frontiers showed similar diminishing-return structure. This reproducibility suggests that the size--error frontier reflects a robust property of the explored reservoir design space for the Kuramoto--Sivashinsky task, rather than a run-specific accident of initialization.

Taken together, these results refine the interpretation of reservoir complexity. Improved prediction was accompanied not by a monotonic reduction in reservoir size, but by the emergence of an empirical efficiency frontier. The relevant analogy is therefore not that evolution makes networks smaller, but that adaptive search organizes computational systems under structural constraints: predictive function emerged through the coordinated tuning of capacity and dynamics, rather than through raw expansion of network size alone.

\subsection{Evolution refines the low-eigenvalue spectrum within an SBM-like envelope}
\label{sec:spectral_signature}

To investigate the structural basis of these evolutionary performance gains, we analyzed the spectrum of the random-walk normalized Laplacian \(\mathbf{L}_{\mathrm{rw}} = \mathbf{I} - \mathbf{D}^{-1}\mathbf{A}\) of the directed weighted reservoir matrix. For each evolved reservoir, the smoothed spectral density \(\Gamma(\lambda)\) was estimated by Gaussian kernel density estimation on the real parts of the eigenvalues of \(\mathbf{L}_{\mathrm{rw}}\) (See Methods sec.~\ref{sec:methods-laplacian}).

\subsubsection{Reservoirs occupy an SBM-like spectral signature class}

\begin{figure}[htbp]
    \centering

    \includegraphics[width=\columnwidth]{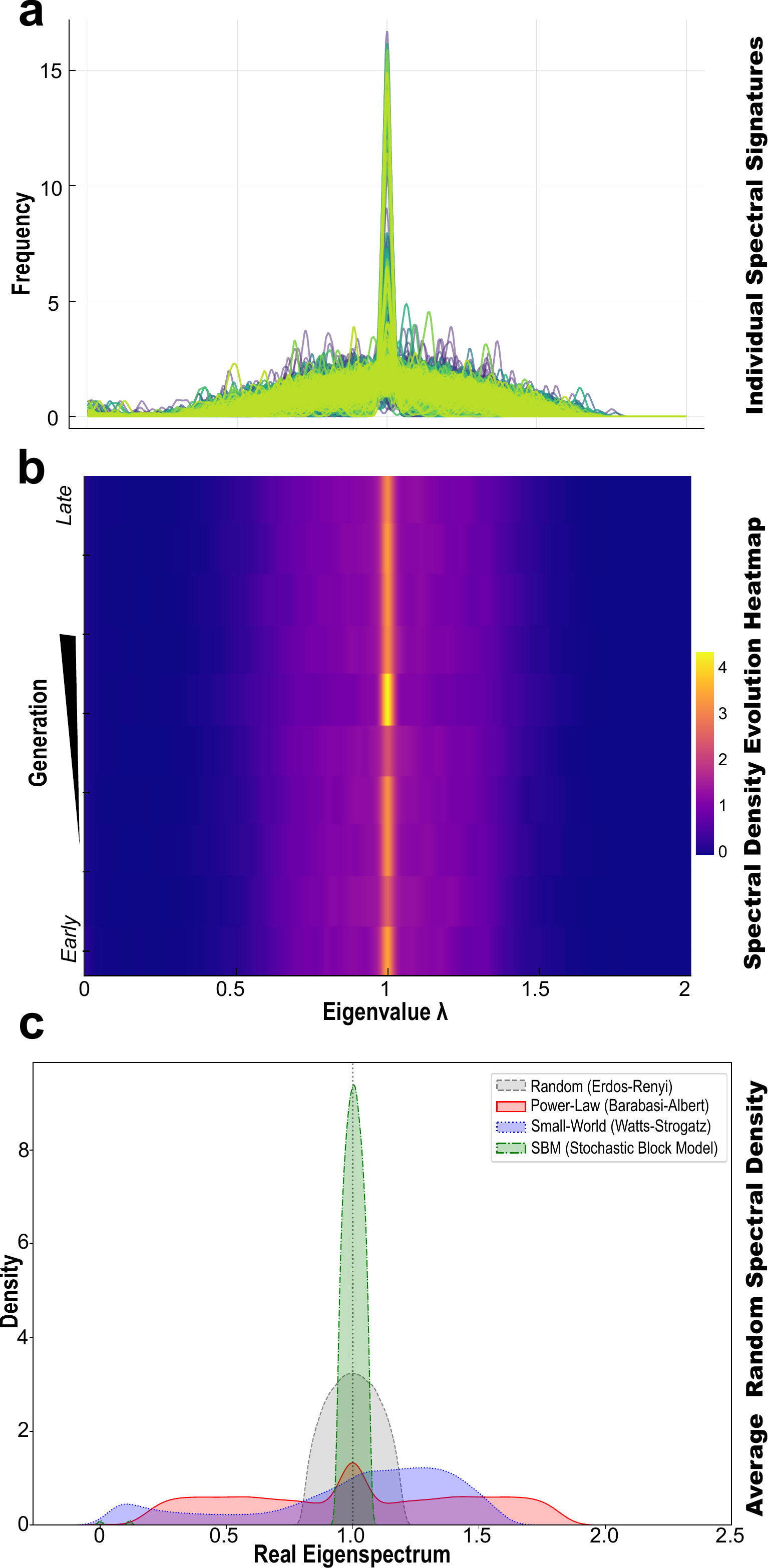}

    \caption{\textbf{Reservoirs occupy an SBM-like spectral signature class within which evolution operates.}
    \textbf{(a)} Smoothed spectral density \(\Gamma(\lambda)\) of the random-walk normalized Laplacian for stratified selection of 800 individuals from ten evenly spaced generations from early (\texttt{Gen~0}) to late (\texttt{Gen~60}).
    \textbf{(b)} Population smoothed spectral density \(\Gamma(\lambda)\) of the random-walk normalized Laplacian, shown at ten evenly spaced generations from~0 to~60 (\texttt{RUN \#8}). The sharp peak at \(\lambda = 1\) and the surrounding bulk are essentially conserved across evolution.
    \textbf{(c)} Comparison against canonical network models (averaged over 100 instantiations for each model). Synthetic baselines were generated as binary graphs (ER: Erd\H{o}s--R\'enyi (ER), Barab\'asi--Albert: BA, Watts--Strogatz: WS, and SBM undirected); the comparison with the directed weighted reservoir spectrum is qualitative (shape only). Only the SBM among these baselines produces the sharp peak at \(\lambda = 1\), the diagnostic feature of community structure---placing the evolved reservoirs in an SBM-like signature spectral class throughout evolution.}
        \label{fig:spectral_comparison}
\end{figure}

Evolved reservoirs exhibit a spectral signature dominated by a sharp peak at \(\lambda = 1.0\) embedded within a broader bulk distribution (Fig.~\ref{fig:spectral_comparison}.a, Fig.~\ref{fig:spectral_comparison}.b). To classify this signature, we compared the reservoir spectrum against synthetic Erd\H{o}s--R\'enyi (ER), Barab\'asi--Albert (BA), Watts--Strogatz, and Stochastic Block Model (SBM) graphs (Fig.~\ref{fig:spectral_comparison}.c). Among these baselines, only the SBM produces the sharp \(\lambda = 1\) spike---the diagnostic spectral feature of structural symmetries and motif redundancy within community blocks. The reservoir spectrum is therefore SBM-like in its peak structure, embedded within a broader bulk spanning roughly \(\lambda \in [0.5,1.5]\) than the canonical SBM and unlike either the smooth ER bulk or the flat BA and Watts--Strogatz distributions.

To a degree, this SBM-like envelope is a property of the reservoir design space rather than a target reached by evolution. The reservoir initialization scheme produces directed, weighted, sparse matrices via random connectivity with i.i.d.\ uniform weights (Methods); the \(\lambda = 1\) peak is a generic spectral feature of the random-walk Laplacian on such matrices, distinct from the binary-graph ER baseline shown in Fig.~\ref{fig:spectral_comparison}.c. The genetic algorithm operates not on the rule for generating the individual matrix entries but on the five hyperparameters that control reservoir construction (size, connectivity degree, spectral radius, input scaling, regularization--See  Fig.~\ref{fig:overview}). Within this design space, the SBM-like spectral signature class is conserved across generations: the population-averaged spectrum at late generation (\texttt{Generation~60}) is essentially indistinguishable from that at early generation (\texttt{Generation~0}). Quantitatively, the spectral centroid shifts by \(0.65\%\) (\(0.983 \to 0.989\)), the peak location remains at \(\lambda = 1.000\), the peak density slightly broadens (\(0.0035 \to 0.0032\)), and the bulk skewness moves from \(-0.415\) at Generation~0 to \(-0.272\) by Generation~60, with most of the skewness change occurring between Generations~0 and~6.

Importantly, note that the observed SBM-like spectral envelope should be interpreted as a spectral resemblance, not as evidence that the reservoirs were initialized from, or exactly recovered, a stochastic block model. The recurrent matrices were initialized as sparse random directed weighted graphs without an imposed block partition. The block structure in the model enters through the input coupling matrix, whereas the Laplacian spectra were computed from the recurrent matrix alone. The persistence of a broad spectral envelope across generations therefore reflects the sparse recurrent ensemble available to evolution, while the systematic changes within that envelope identify the subset of recurrent architectures preferentially selected for prediction.

The conservation of the global spectral class is itself informative. It establishes that evolution operates \emph{within} a structurally constrained envelope rather than transforming the reservoir into a different architectural family. The substantive question is then where, within this envelope, evolution exerts directional pressure---and the next analyses identify a specific spectral regime that is targeted by selection.

\subsubsection{Asymmetric selection pressure targets the low-eigenvalue end of the spectrum}

While the global spectral envelope is conserved, evolution exerts pronounced and asymmetric selection pressure on specific spectral regimes. K-means clustering of the smallest two eigenvalues \((\lambda_1, \lambda_2)\) reveals three distinct archetypes that occupy separable regions of the low-frequency plane (Fig.~\ref{fig:clustering_emd}.a). The same scatter, recolored by generation, shows a clear directional trajectory: early-generation reservoirs (light shading) populate the high-\(\lambda_1\) tail, extending out to \(\lambda_1 \approx 0.25\), while late-generation reservoirs (dark shading) condense progressively toward the \(\lambda_1 \to 0\) axis. The genetic algorithm thus drives the population away from the high-\(\lambda_1\) regime---which corresponds to faster algebraic relaxation and shorter integration timescales---and toward the low-\(\lambda_1\) regime that supports longer reservoir memory.

This selection pressure is sharply localized at the bottom of the spectrum. K-means clustering applied to the high-eigenvalue end shows no comparable generational drift: the pair \((\lambda_{\max}, \lambda_{\max-1})\) lies along a tight collinear ridge with no separable cluster structure, and the projection onto \((\lambda_{\max}, \lambda_{\max-2})\) reveals three clusters that are populated indistinguishably by early- and late-generation reservoirs (Fig.~\ref{fig:clustering_emd}.b). The high-frequency oscillatory modes of the reservoir are thus structurally indifferent to selection, while the long-timescale modes at the bottom of the spectrum are the direct targets of evolutionary optimization.

Principal component analysis on the full 200-dimensional spectrum confirms this localization and identifies the spectral mode that tracks predictive performance. The first two principal components capture 35\% and 24\% of the population spectral variance respectively, but neither correlates with prediction error \(J\) (\(|r| < 0.07\) for both). The third principal component, despite explaining only 3.4\% of the variance, loads heavily on the smallest eigenvalues (positions \(\lambda \in [0.005, 0.02]\)) and correlates strongly with performance (\(r = 0.54\)). The dominant axes of population variation (PC1 and PC2, distinguishing between archetypes) thus do not predict performance; rather, the predictive signal lives in a low-variance mode at the very bottom of the spectrum---precisely where the K-means analysis localized the generational drift. Evolution therefore does not just modify the spectrum somewhere; it modifies the specific spectral region that controls predictive function.

\subsubsection{Spectral conservation and selective refinement coexist}

\begin{figure}[hb!]
    \centering
    \includegraphics[width=\columnwidth]{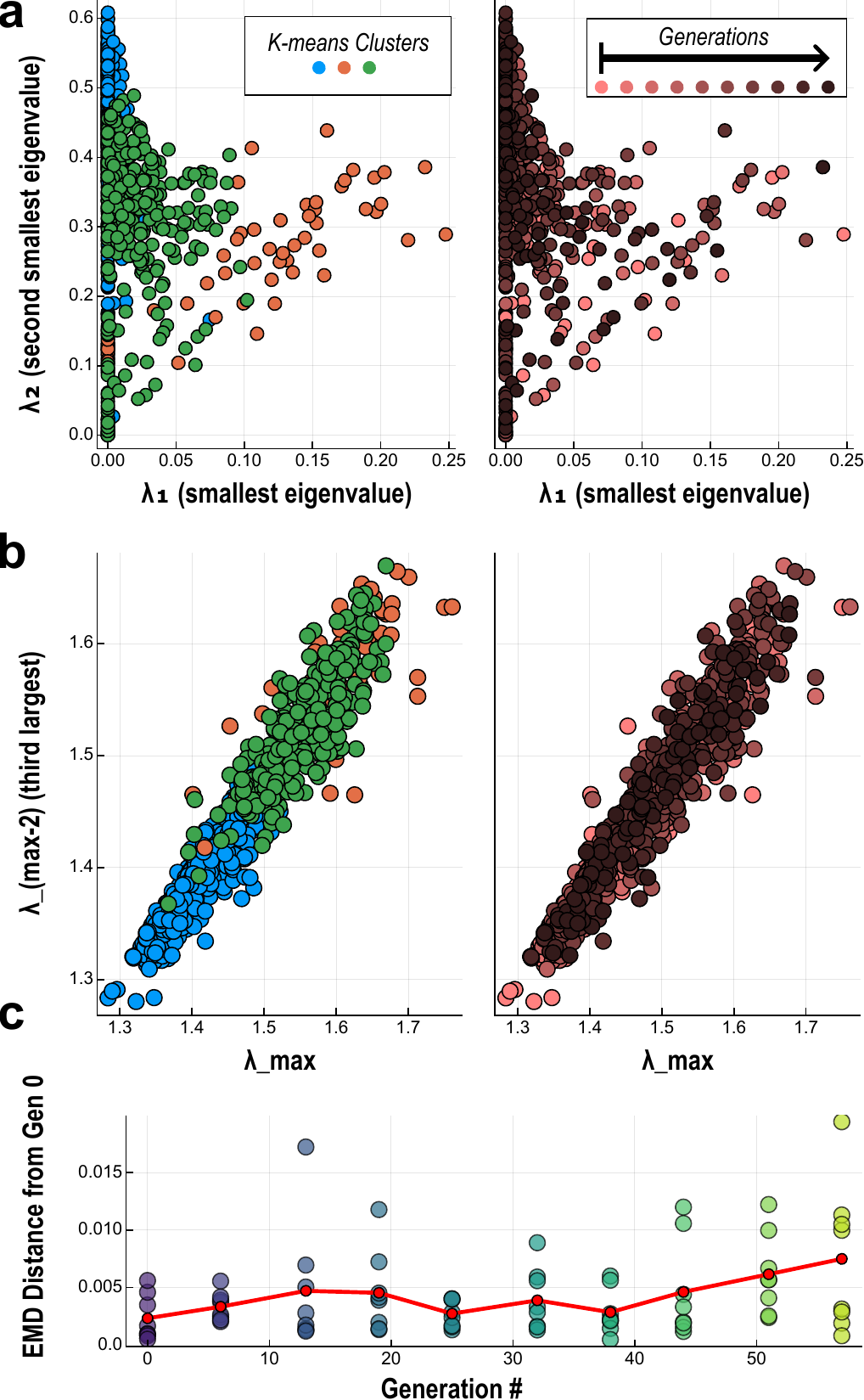}
    \caption{\textbf{Asymmetric selection pressure targets the low-eigenvalue spectrum.}
    \textbf{(a)} K-means clustering of the smallest two eigenvalues \((\lambda_1, \lambda_2)\). Left: three identified clusters (color denotes cluster identity). Right: same scatter colored by generation (light = early, dark = late). The high-\(\lambda_1\) tail is populated almost exclusively by early generations; later generations condense toward the \(\lambda_1 \to 0\) axis, indicating directional drift toward longer memory timescales.
    \textbf{(b)} K-means clustering of \((\lambda_{\max}, \lambda_{\max-2})\). Left: three clusters in this projection. Right: same scatter colored by generation, showing no systematic temporal pattern. The high-frequency end of the spectrum is structurally indifferent to selection. The pair \((\lambda_{\max}, \lambda_{\max-1})\) (not shown) is collinear with no separable cluster structure, motivating the use of \(\lambda_{\max-2}\).
    \textbf{(c)} Earth Mover's Distance (EMD) from the Generation~0 representative pool, plotted against generation. Mean EMD (red) increases modestly across evolution, with absolute values remaining small. EMD distance vs.\ prediction error \(J\) (\(r = 0.413\)). Reservoirs with larger spectral drift tend to predict worse, indicating that evolution stabilizes the global spectral template while exerting targeted refinement at specific spectral regimes.}
    \label{fig:clustering_emd}
\end{figure}

To quantify the cumulative spectral reorganization across evolution, we computed the Earth Mover's Distance (EMD) between each evolved spectrum and the Generation~0 representative pool. The mean EMD increases modestly with generation (early-generation average \(\approx 0.0035\); late-generation average \(\approx 0.005\); Fig.~\ref{fig:clustering_emd}.c), consistent with directional drift away from the initialization spectrum. The absolute drift is small in magnitude, reflecting the conservation of the global spectral class identified in Fig.~\ref{fig:spectral_comparison}.

Across the population, EMD distance from \texttt{Generation~0} is positively correlated with prediction error (\(r = 0.413\). Reservoirs with the largest spectral drift from the initialization template tend to perform worse, not better. Combined with the K-means and PCA results, this defines the geometry of evolutionary action in spectral space: gross drift away from the SBM-like envelope is selected against, while targeted refinement at the low-eigenvalue end is selected for. The spectral envelope and the bottom-of-spectrum mode are simultaneously under selection, but in opposite directions---the envelope is stabilized while the low-\(\lambda\) regime is directionally pushed toward longer memory timescales.

This combination of conservation and directional refinement parallels the structural pattern we will see in the macroscopic analyses: modularity is held fixed within a sharp band while connection cost is actively pruned, and elite predictive networks occupy a horizontal Pareto floor at the joint structural minimum (Sec.~\ref{sec:multiobj}). Evolution thus operates with the same logic across spectral and structural axes: conserve the global organizing template, refine the specific feature that matters for function.

\subsection{Evolutionary pressure incentivizes modularity}

\subsubsection{Empirical convergence to an intermediate modularity}
\label{sec:modularity_convergence}

The spectral and EMD analyses indicated that evolution operates within a conserved spectral envelope, with refinement concentrated at the low-eigenvalue end of the spectrum. To complement this spectral characterization with a direct macroscopic measure of network structure, we analyzed the distribution of Newman modularity \(Q\) and connection cost across the evolved reservoir population. Modularity was computed from the directed weighted reservoir matrix using label-propagation community detection (Methods); both modularity and connection cost were min-max normalized to \([0,1]\) for comparison.

\begin{figure}[htbp]
    \centering
    \includegraphics[width=\columnwidth]{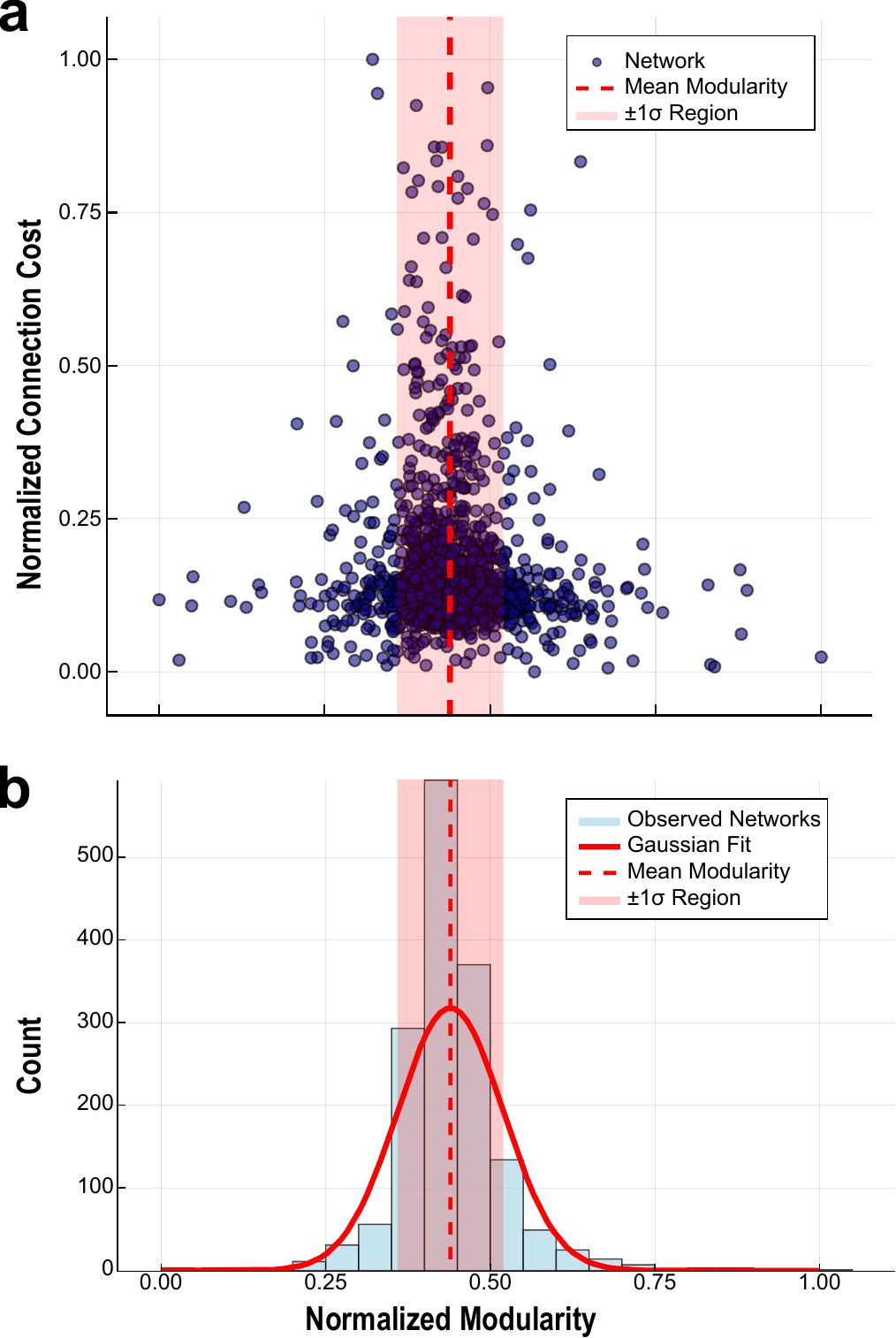}
        \caption{\textbf{Empirical convergence to an intermediate macroscopic modularity.}
        (a) Joint distribution of normalized connection cost (vertical) and normalized modularity (horizontal) across the evolved reservoir population. Each marker is one reservoir, sampled across 20 generations using the performance-stratified strategy described in Methods. Marker color encodes prediction error \(J\). The vertical dashed line and shaded band mark the bell-curve peak \(\mu = 0.439\) and the \(\pm 1\sigma\) region from the Gaussian fit in panel (b). The population forms a funnel: at high cost, reservoirs span a broad range of modularity values; at low cost, modularity is narrowly concentrated near \(\mu \approx 0.44\). A small isolated cluster at low modularity / intermediate cost corresponds to a distinct minority of outlier reservoirs.
        (b) Marginal distribution of normalized modularity across the same sample. A Gaussian fit (red curve) returns \(\mu = 0.439\) and \(\sigma = 0.08\). The empirical histogram is leptokurtic relative to the fit, with a sharp central spike (peak sharpness \(\approx 12.4\)). The shaded band marks the \(\pm 1\sigma\) region.}
    \label{fig:modularity_bell_curve}
\end{figure}

In the joint cost--modularity plane, the evolved population forms a characteristic funnel shape (Fig.~\ref{fig:modularity_bell_curve}.a). At high connection cost, reservoirs span a broad range of modularity values. As cost decreases, the tolerated range of modularity narrows, with the low-cost reservoirs concentrating in a tight vertical band centered at \(\mu \approx 0.44\) on the normalized modularity axis. The marginal distribution of normalized modularity, aggregated across all sampled generations, is sharply peaked at this central value (Fig.~\ref{fig:modularity_bell_curve}.b). A Gaussian fit to the histogram returns \(\mu = 0.439\) and \(\sigma = 0.08\), but the empirical distribution is distinctly leptokurtic relative to this fit: the network counts overshoot the Gaussian model at the center, producing a sharply peaked profile (peak sharpness \(\approx 12.4\)). The bulk of the evolved population is contained within the \(\pm 1\sigma\) band of this central peak.

We note that population-level linear correlations between modularity and prediction error \(J\) are weak (\(r = 0.08\)), and the same is true for connection cost (\(r = -0.07\)). The funnel geometry is therefore not captured by simple linear regression at the population level; rather, it reflects a conditional structure that becomes apparent only in the low-cost regime, where modularity is increasingly constrained. This pattern is consistent with stabilizing selection on a topological constraint that becomes binding as connection cost is reduced, rather than with a single modularity value being directly selected for higher performance. This pattern is the signature of stabilizing selection: successful prediction of the target spatiotemporal dynamics required a precise intermediate balance of structural segregation and integration, rather than maximization of either property.

\subsubsection{Temporal dynamics of structural pruning under topological constraint}
\label{sec:temporal_dynamics}

\begin{figure}[htbp]
    \centering
        \centering
        \includegraphics[width=\columnwidth]{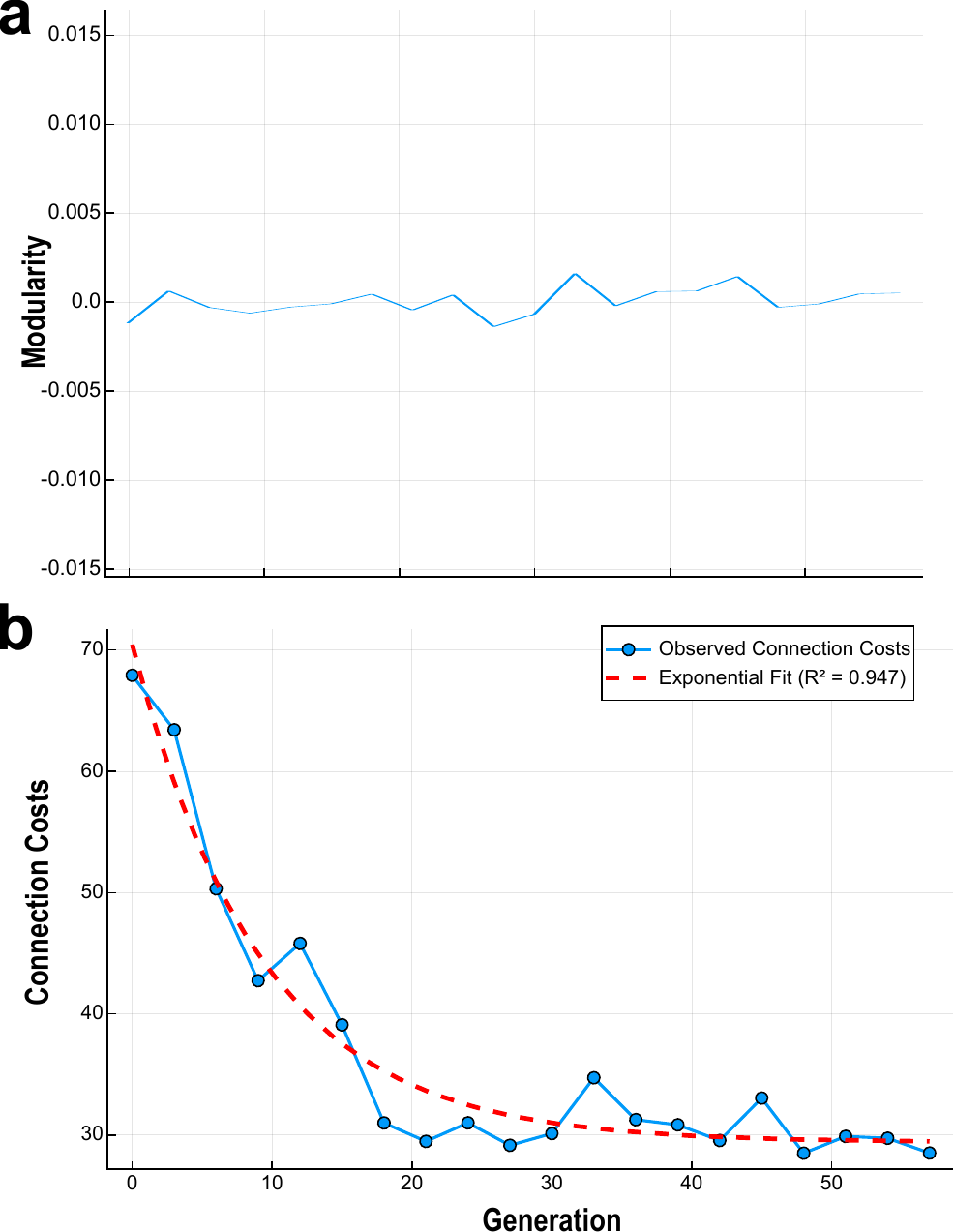}
    \caption{\textbf{Connection cost is reduced exponentially while modularity remains stable.}
    \textbf{(a)} Population-averaged raw Newman modularity \(Q\) over the same generations. Per-generation means fluctuate within a narrow strip of width \(\pm 0.002\) near zero, occupying a small fraction of the full population range. The intermediate modularity is maintained throughout the run rather than approached gradually, consistent with stabilizing selection on macroscopic modularity.
    \textbf{(b)} Population-averaged connection cost as a function of generation. Blue markers, observed data; dashed red curve, exponential fit \(f(g) = 41.07\,e^{-0.108\,g} + 29.39\) (\(R^2 = 0.95\), half-life \(\approx 6.4\) generations).    
    }
    \label{fig:temporal_dynamics}
\end{figure}

Having established that the evolved population concentrates in a narrow modularity band, we examined how this structural target is expressed over evolutionary time. The two structural axes show contrasting trajectories: connection cost decreases rapidly and continuously across generations, whereas modularity remains essentially fixed throughout (Fig.~\ref{fig:temporal_dynamics}).

The population-averaged connection cost decreases with generation along an exponential decay (Fig.~\ref{fig:temporal_dynamics}.b; \(f(g) = 41.07\,e^{-0.108\,g} + 29.39\), \(R^2 = 0.95\), half-life \(\approx 6.4\) generations), dropping from an initial average of \(\sim 68\) to a stable asymptote near \(\sim 29\). The randomly initialized reservoirs are thus over-wired relative to the requirements of Kuramoto--Sivashinsky prediction, and the GA prunes this excess recurrent connectivity rapidly during the early generations before settling into a low-cost regime.

The same population, over the same generations, exhibits no comparable change in modularity (Fig.~\ref{fig:temporal_dynamics}.a). Whereas connection cost is reduced by more than half, the per-generation mean of raw Newman \(Q\) fluctuates within a narrow strip near zero (\(\pm 0.002\)), occupying only a small fraction of the full population range \([-0.028, 0.036]\) observed across reservoirs. The intermediate modularity target identified in Fig.~\ref{fig:modularity_bell_curve} is therefore not approached gradually; it is present from the earliest generations and maintained as the population is pruned. This pattern is consistent with stabilizing selection on modularity: the topological balance is preserved as a constraint, while structural cost is the variable that evolution actively reduces within that constraint. The optimization therefore does not appear to trade modularity against cost continuously; instead, it confines reservoirs to a narrow modularity regime while pruning excess recurrent connectivity within that regime.

\subsubsection{Multi-objective Pareto analysis confirms the topological constraint}
\label{sec:multiobj}
The size--error analysis in Sec.~\ref{sec:size_efficiency} characterized the evolved reservoir population descriptively, by dominance sorting of the evolved data. To complement this with a prescriptive analysis, we applied a post hoc multi-objective optimization---the NSGA-II framework---as an analytical tool to quantify the joint trade-off among prediction error, connection cost, and modularity. For clarity, note that NSGA-II is used here not to evolve new reservoirs but to compute theoretical Pareto optima in the normalized cost--modularity plane, against which the empirical population can be benchmarked. Four composite objective functions were constructed to balance performance against connection cost and modularity (For details, see Methods.~\ref{sec:method-nsga}). Briefly, the four composite objectives capture distinct trade-offs in normalized metric space. \(O_1\) measures \emph{improvement across generations} by rewarding low normalized prediction error while down-weighting later generations: \(O_1=\mathrm{norm\_performance}/(1+\mathrm{norm\_generation})\). \(O_2\) encodes \emph{structural efficiency}, favoring networks that minimize wiring cost for a given modular organization: \(O_2=\mathrm{norm\_connection\_cost}/(1+\mathrm{norm\_modularity})\). \(O_3\) probes the \emph{performance–modularity relationship} by penalizing error in proportion to modularity: \(O_3=\mathrm{norm\_performance}/(1+\mathrm{norm\_modularity})\). Finally, \(O_4\) captures the \emph{performance–cost trade-off} by amplifying prediction error when connection cost is high: \(O_4=\mathrm{norm\_performance}\,(1+\mathrm{norm\_connection\_cost})\). These composites were optimized with NSGA-II to produce theoretical Pareto fronts against which the empirical population was compared.

\begin{figure}[htbp]
    \centering

        \includegraphics[width=\columnwidth]{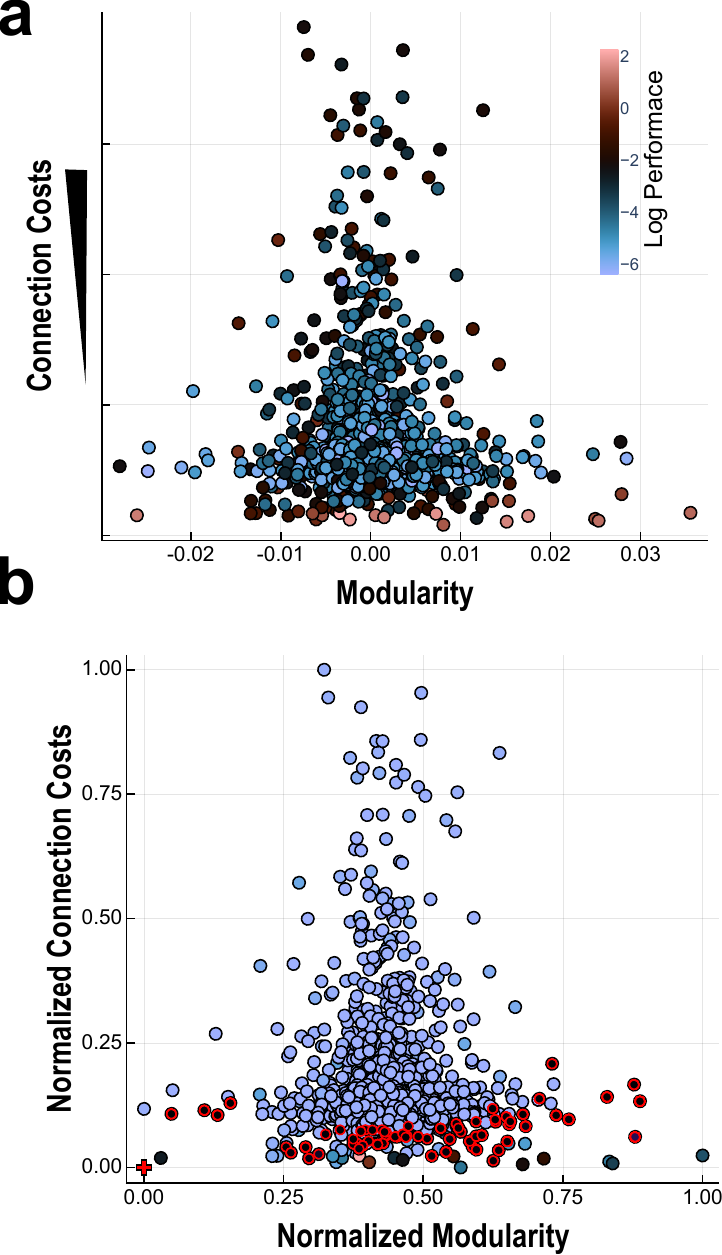}

    \caption{\textbf{Multi-objective Pareto analysis of the cost--modularity--performance trade-off.}
    \textbf{(a)} Distribution of evolved reservoirs in the raw cost--modularity plane (raw connection cost on the vertical axis; raw Newman \(Q\) on the horizontal axis). Color encodes \(\log\)-scaled prediction error \(J\). The data form a funnel: high-cost networks span a broad range of modularity, while low-cost, low-error networks are confined to a narrow modularity band around weak raw Newman \(Q\) values, corresponding to \(\mu \approx 0.44\) on the normalized scale. 
    \textbf{(b)} NSGA-II multi-objective Pareto analysis on normalized axes. Blue points: full evolved population in normalized cost--modularity space. Red points: theoretical Pareto-optimal solutions returned by NSGA-II and their closest empirical matches. The empirical Pareto-optimal points form a horizontal band at the floor of normalized cost rather than a sloped trade-off curve, indicating that elite predictive performance is achieved by minimizing connection cost \emph{at} the constrained modularity, rather than by trading one against the other. Note the change of axis between panels: panel (a) is on raw scales (suitable for connecting to the per-network modularity values); panel (b) is on normalized scales (required for NSGA-II in objective space).}
    \label{fig:pareto_frontier}
\end{figure}

The empirical distribution of evolved reservoirs in the cost--modularity plane shows the same funnel shape on raw axes that was visible in the normalized landscape of Fig.~\ref{fig:modularity_bell_curve}a, (Fig.~\ref{fig:pareto_frontier}a). At high connection cost, reservoirs occupy a wide range of modularity values and produce a broad spread of prediction errors. As cost is reduced---moving downward along the cost axis---the tolerated range of modularity narrows, and the low-error reservoirs (darker color) concentrate in a tight band along mean \(Q \approx 0\). Efficiency and accuracy are therefore not freely combinable across the modularity axis: the empirical population obeys a topological boundary condition that becomes increasingly tight as cost decreases.

The NSGA-II multiobjective analysis quantifies this boundary (Fig.~\ref{fig:pareto_frontier}b). From a population of 1000 candidate solutions in normalized cost-modularity space, the algorithm returned theoretical Pareto-optimal points that lie at the joint floor of normalized cost (\(\to 0\)). Mapping each theoretical optimum (of multiobjective optimization of the 4 functions) to its closest empirical match in the evolved population shows strong convergence: the median Euclidean distance between empirical reservoirs and the theoretical Pareto frontier was \(0.137\), with 220 networks within \(10\%\) and 1{,}273 of 1{,}600 within \(20\%\) of the frontier.

The geometry of the empirical Pareto frontier is informative: rather than tracing a sloped trade-off curve between cost and modularity, the empirical Pareto-optimal points form a horizontal band that hugs the floor of permissible connection cost while spanning the modularity axis. Elite predictive performance therefore does not require a compromise between cost and modularity; instead, it requires that connection cost be minimized within the modularity band that the rest of the evolved population also occupies. 

\section{Discussion}

We evolved reservoir computing networks for prediction of Kuramoto--Sivashinsky dynamics and asked how predictive function relates to the structural and dynamical organization of the recurrent substrate. The genetic algorithm reduced the composite prediction-error score \(J\) systematically across generations, but the form of this reduction is as important as its magnitude. Improvement was not restricted to a small elite subset of reservoirs while the rest of the population stagnated. Instead, the entire distribution shifted toward lower error, high-error reservoirs were progressively eliminated, and the spread of \(J\) narrowed across generations. In a reservoir-computing setting where each generation samples stochastic reservoirs from a favored hyperparameter regime, this indicates that the GA identified regions of parameter space in which the \emph{typical} reservoir, not only the exceptional one, is well matched to KS prediction. The reduction in population variance is therefore as informative as the reduction in the mean. 

Although evolutionary and metaheuristic methods have often been used to improve reservoir or neural-network performance \cite{clune2013evolutionaryorigins,dale2018neuroevolutionofhierarchical,ferreira2009geneticalgorithmfor,Schmidhuber2007evo,basterrech2023revisitingreservoircomputing,Stanley2019neuroevo}, less emphasis has been placed on using the evolved population itself to infer structure--function constraints. Here, evolutionary optimization is used not only as a search procedure, but also as a mechanistic probe of recurrent dynamical substrates. By tracking prediction error together with forecast horizon, size, spectra, modularity, and connection cost across generations, we identify which architectural degrees of freedom become constrained by successful prediction of a spatially extended chaotic system.

The improvement in \(J\) corresponds to a dynamical change in forecast horizon. A scalar score could improve in several ways: by reducing only early prediction error, by distributing moderate error across the prediction window, or by preserving low error for longer intervals before chaotic divergence accumulates. The time-resolved NRMSE analysis distinguishes among these possibilities and shows that evolution favors the third. Evolved reservoirs maintain $\mathrm{NRMSE}<0.05$ for longer portions of the prediction window, indicating that the optimized recurrent substrates remain dynamically aligned with the target system for extended finite horizons. This is central for chaotic prediction, where small discrepancies grow rapidly under sensitive dependence on initial conditions and performance is naturally measured over valid prediction windows or Lyapunov-scaled times \cite{pathak2018modelfreepredictionof,fleddermann2025improvingtheprediction}. The reduction in \(J\) is therefore not merely an aggregate error effect; it reflects an extension of dynamically meaningful prediction.

The empirical size--efficiency frontier provides the structural counterpart of this result. Reservoir size could have been selected monotonically, with larger reservoirs always favored because they provide more degrees of freedom, or smaller reservoirs favored because they are cheaper and less prone to irrelevant modes. We find neither outcome. Instead, the population condenses around a diminishing-return frontier: increasing reservoir size can reduce attainable error along the nondominated boundary, but with rapidly decreasing marginal gains, and many large reservoirs remain dominated when their other hyperparameters are poorly tuned. This agrees with the broader observation that reservoir performance is not determined by size or spectral radius alone, but by the joint tuning of architecture, connectivity, scaling, and task-dependent dynamical regime \cite{ozturk2007analysisanddesign,roeschies2010structureoptimizationof}. Size is therefore a permissive condition rather than a sufficient one. A reservoir must be large enough to support the dynamical complexity required by KS, but additional capacity has limited value unless spectral radius, connectivity degree, input scaling, and regularization are jointly tuned.

Together, these findings---population-level reduction in \(J\), extension of the low-error forecast horizon, and the diminishing-return size--efficiency frontier---reframe what evolutionary optimization is doing to the reservoir. It is not simply preserving isolated high-performing networks. It is reorganizing the population around a coordinated hyperparameter regime whose typical realizations match the dynamical demands of the target system. The structural analyses then ask what this match looks like at the level of network spectra, modularity, and cost.

\subsection{Conservation and refinement are simultaneous, and the conserved class is task-shaped}

The clearest pattern across our analyses is that evolution exerts two distinct forces at once: it stabilizes the global organizing template and it directionally refines a specific feature within that template. At the spectral level, an SBM-like envelope is preserved across the full evolutionary run; reservoirs that drift away from this envelope predict worse, not better. This interpretation is consistent with previous work showing that normalized Laplacian spectra can reveal conserved architectural classes in neural networks across species and scales, including spectral peaks near \(\lambda=1\) and low-eigenvalue structure associated with community organization \cite{delange2014neuralspectrum}. Within the envelope, the bottom of the spectrum is actively pushed toward \(\lambda_1 \to 0\), and a low-variance principal component (PC3, \(3.4\%\) of population spectral variance) loading on the smallest eigenvalues is the only spectral mode that correlates substantially with prediction error. At the structural level, modularity is held within a sharp band (\(\sigma = 0.08\) on the normalized scale) while connection cost decays exponentially with a half-life of six generations; in the joint cost--modularity plane, elite predictive networks occupy a horizontal Pareto floor rather than a sloped trade-off curve.

Why does evolution preserve an SBM-like envelope rather than, say, an Erd\H{o}s--R\'enyi or scale-free one? This question is non-trivial because the reservoir initialization scheme generates Erd\H{o}s--R\'enyi-like topology: the construction routine produces directed sparse matrices with i.i.d.\ uniform weights on a random support, with no community structure imposed by design. Prior work has shown that reservoir topology can influence memory, prediction, and task performance, and that modular, small-world, empirical-connectome, or stochastic-block-like structures can alter reservoir dynamics relative to homogeneous random networks \cite{carroll2019networkstructure,haluszczynski2019goodandbad,kitayama2022guidingprincipleof,mcdaniel2022investigatingechostate,yang2024braininspiredmodularecho,damicelli2022brainconnectivitymeets}. The SBM-like spectral signature in our reservoirs emerges from the joint statistics of the favored hyperparameter regime, not from the construction primitive. The relevant question is therefore: which features of the Kuramoto--Sivashinsky target make an SBM-like recurrent substrate the favored one?

The Kuramoto--Sivashinsky equation produces spatiotemporal chaos with a characteristic spatial correlation length set by the most unstable wavenumber and a characteristic temporal correlation time set by the leading Lyapunov exponent \cite{rost1995aparticlemodel,tomlin2018nonlineardynamicsof,pathak2018modelfreepredictionof}. Predicting such dynamics requires a reservoir whose internal state can simultaneously hold spatially distinct modes, so that distant grid points are not represented redundantly, and temporally coherent dynamics, so that information is integrated across the system's correlation time. A network with too little community structure---an Erd\H{o}s--R\'enyi-like substrate with uniform mixing---fails the spatial requirement: every reservoir unit becomes effectively similar in its statistical relationship to every input channel, collapsing the available dynamical degrees of freedom. A network with too much community structure---rigid, weakly connected blocks---fails the temporal requirement: information cannot integrate across modules within the relevant timescale. The SBM-like  organization that evolution favors sits between these extremes: enough community structure to support distinct functional roles for different parts of the network, but enough between-module connectivity to permit integration. This interpretation is consistent with the broader network-science view that modular organization supports segregation and specialization, while connector structure and inter-module pathways support integration \cite{sporns2016modularbrainnetworks,meunier2010modularandhierarchically,Chen2021modularFrontalCortex,Ji2015modularVisualCortex}. The class is conserved because it is the right class for this task; if the task changes, the conserved class should change with it.

This framing also explains why the directional refinement is so sharply localized at the bottom of the spectrum. Once the reservoir has been pushed into the SBM-like spectral signature class with broadly appropriate community structure, the remaining functional question is timescale: does the reservoir's slowest mode match the target system's correlation time? Small eigenvalues of the normalized Laplacian are closely tied to connected components, slow graph modes, and community structure, making the low-\(\lambda\) regime a natural place to look for structure-function relationships in recurrent networks \cite{delange2014neuralspectrum,mcgraw2008laplacianspectraas}.

In echo state networks, spectral radius and eigenvalue distributions shape reservoir memory, effective timescales, and operating regime; spectral radius alone is an incomplete descriptor of the dynamical capacities relevant for prediction \cite{ozturk2007analysisanddesign,verstraeten2009onthequantification,caluwaerts2013thespectralradius,li2019effectsofsingular,carroll2020edgeofchaos}. This is the variable that selection actively tunes. The high end of the spectrum, governing fast responses, is left structurally indifferent because any sufficiently fast response is acceptable. Long timescales must be tuned; short timescales need only exist.

The reason the predictive signal lives in PC3 rather than PC1 or PC2 follows from this same logic. PC1 and PC2 capture the dominant axes along which reservoirs differ from one another within the conserved spectral envelope---they index variation that selection has already deemed permissible. PC3, despite explaining only \(3.4\%\) of population variance, captures the residual variation in the low-\(\lambda\) regime that selection has not yet equalized, and it is this residual variation that modulates predictive function. A naive analysis based on dominant axes of population variation would have missed the relevant dimension entirely. The lesson is that in evolutionary systems where a global architectural class is preserved, function is encoded in low-variance modes hidden inside a much larger envelope of conserved structure. The fact that PC3 loads on the same spectral region (\(\lambda \in [0.005,0.02]\)) where K-means clustering localizes the generational drift is the strongest single piece of evidence that what selection is acting on, and what predicts performance, are the same thing.


This is not a description of evolution moving sequentially through stages of exploration and exploitation, nor of the GA discovering a single global optimum. It is a description of evolution operating with two scales of action superimposed on the same population: stabilizing selection on the template, directional selection on the feature. The two scales target different statistical properties of the population. Stabilizing selection compresses variance around the modal architecture: the SBM-like spike, the modularity peak, and the conserved population-averaged spectrum. Directional selection moves the mean of a specific marginal distribution, such as \(\lambda_1\) or connection cost, toward functionally favorable values without disrupting the surrounding template.

The reason this distinction matters is that the predictive signal lives in the directionally selected feature, not in the stabilized template. PC1 and PC2 capture most of the population spectral variance and have negligible correlation with performance. PC3 captures very little variance and has a strong correlation with performance. A naive analysis based on dominant axes of population variation would have missed the relevant dimension entirely. The lesson is that in evolutionary systems where a global architectural class is preserved by selection, function may be encoded in a small-variance mode hidden inside a much larger envelope of conserved structure---and that this is precisely the geometry that selection produces, not an analytical accident.



\subsection{What the GA actually selects on, and why it matters}

A subtle but important methodological point shapes the interpretation of every structural finding: the genetic algorithm did not directly mutate entries of the reservoir connectivity matrix. The matrix is regenerated stochastically for each individual via a random-construction procedure with i.i.d.\ uniform weights on a sparse directed support. The GA mutates and selects on the five hyperparameters that parameterize this construction: size, connectivity degree, spectral radius, input scaling, and regularization. The structural patterns we observe---the size--efficiency frontier, the spectral refinement at low \(\lambda\), the modularity band, and the exponential cost decay---are population-level fingerprints of which hyperparameter regimes get favored, not sculpted features of individual matrices.

This is not a limitation; it is a closer analogue of how genes encode neural circuits than direct connectome mutation would be. The mammalian brain is not specified synapse-by-synapse in the genome. Instead, genes specify developmental programs, molecular cues, cell-type identities, growth constraints, and probabilistic rules from which circuit structure emerges with statistical regularities \cite{mitchell2024variabilityinneural,luo2021architecturesofneuronal,barabasi2020ageneticmodel}. Recent generative and constructive-connectomics perspectives make this point in explicit computational terms: compact developmental rules can generate structured but variable connectomes without requiring a literal connection table \cite{richter2025buildingtheconnectome,stan2022constructiveconnectomicshow}. The evolved reservoirs in this study are subject to an analogous logic: a small set of construction parameters is under selection, and structural regularities such as modularity, spectra with suppressed high-\(\lambda_1\) tails, and costs emerge as the joint statistics of that parameter regime.

This framing also explains why the evolved population concentrates so sharply rather than dispersing across the structural space. The narrowness of the modularity band, the leptokurtic peak shape, and the horizontal Pareto floor are not signs that the GA found a single global optimum and then stopped. They are signs that the favored hyperparameter regime is itself narrow, and that random-construction reservoirs sampled from this narrow regime produce structural features with low cross-instance variance. The selection is on a parameter regime; the structural concentration is its consequence. This parameter-regime view also parallels the way evolved biological regulation is often organized around motifs, feedback loops, modular architectures, and higher-level control variables rather than around independent tuning of every microscopic interaction \cite{dehghani2024cellularcomplexity}. This distinction is invisible from the structural data alone but is fundamental to understanding what the GA is doing.

\subsection{Why the bottom of the spectrum is where selection acts}

The localization of selection pressure at the bottom of the eigenvalue distribution is consistent with a specific dynamical interpretation. In recurrent dynamics 
, the smallest non-trivial eigenvalues set the slowest relaxation timescales of the system. A reservoir with \(\lambda_1 \to 0\) has at least one mode whose associated decay time grows very long; this is the regime in which the reservoir can integrate information across long temporal windows. The Kuramoto--Sivashinsky system has a specific Lyapunov spectrum and a specific correlation time, and predicting its evolution requires holding state across that correlation time. The spectral feature that evolution refines---suppression of the high-\(\lambda_1\) tail and concentration near \(\lambda_1 \to 0\)---is precisely the feature that controls whether the reservoir's intrinsic timescales match the timescales of the system it must predict.

The high end of the spectrum, by contrast, controls fast components of the reservoir's response. These fast modes are required for the reservoir to be responsive to inputs, but they do not need to match any specific timescale of the target system as long as they are sufficiently fast. The evolutionary indifference of the high-\(\lambda\) regime in our data is therefore not an absence of selection pressure on fast dynamics; it is selection that does not discriminate among fast modes once they are fast enough. Long timescales must be tuned; short timescales need only exist. This is consistent with the broader principle that control of complex neural systems should be matched to the scale at which the relevant computation is expressed, rather than to the finest available physical scale \cite{dehghani2018multiscalecontrol}.

This asymmetry has a direct counterpart in biological circuits. Cortical and subcortical networks differentiate between fast computational dynamics and slower integrative dynamics that support memory, anticipation, and behavioral state. A large body of work has described hierarchical neural timescales, with faster dynamics in sensory systems and progressively slower dynamics in association and prefrontal circuits, shaped by recurrent connectivity, cortical microarchitecture, and large-scale feedback structure \cite{kiebel2008ahierarchyof,chaudhuri2015alargescalecircuit,gao2020neuronaltimescalesare,raut2020hierarchicaldynamicsas}. The fact that adaptive search through a constrained reservoir design space spontaneously produces a related asymmetry suggests that this may be a generic property of recurrent computation under predictive demands: the slow modes carry the function, and they are what selection acts on.

\subsection{Modularity, cost, and the geometry of the structural Pareto floor}

Two findings about the reservoirs' macroscopic organization are individually interesting and jointly more so. First, modularity locks at an intermediate value, with \(\mu = 0.44\) on the normalized scale, corresponding to weak Newman \(Q\) on the raw scale, and with a sharp leptokurtic peak. This means the population converges to a topological balance between integration and segregation: enough community structure to support distinct functional roles in different parts of the network, but not so much that the network fragments into disconnected modules. The narrowness of this peak indicates that this balance is not a soft preference but a strict requirement. This interpretation is consistent with network accounts of brain organization in which modularity supports specialized local processing, while inter-module connectivity and connector hubs enable integration across distributed systems \cite{sporns2016modularbrainnetworks,meunier2010modularandhierarchically}. Modular organization has also been reported across diverse cortical systems, including early visual and frontal cortex, suggesting that modularity is a recurring principle of cortical organization rather than a feature of a single sensory or cognitive domain \cite{Ji2015modularVisualCortex,Chen2021modularFrontalCortex,Powell2024modularitydiverse}.

This result also connects to evolutionary accounts of modularity. A central idea in this literature is that modular organization can emerge when performance is optimized under pressures that penalize excessive or inefficient connectivity, so that modularity arises as a consequence of economical adaptation rather than as an independently specified design principle \cite{clune2013evolutionaryorigins}. Related work further suggests that modular neural architectures can improve evolvability by allowing systems to acquire new functions while reducing interference with previously acquired ones \cite{ellefsen2015neuralmodularity}. In the present reservoirs, modularity was not directly imposed as an objective, yet the evolved population collapses into a narrow intermediate modularity band. This supports the interpretation that predictive selection favors a constrained balance between segregation and integration.

Second, connection cost is exponentially pruned within the locked modularity band, and elite predictive networks occupy a horizontal floor in the joint cost--modularity plane. The horizontal geometry is significant. A sloped Pareto frontier would indicate that cost and modularity must be traded off against each other; the floor instead indicates that they are independently constrained, with cost minimized within the modularity band rather than against it. Multi-objective and Pareto frameworks are widely used to formalize this kind of constraint geometry, where multiple performance or resource objectives jointly delimit the set of non-dominated solutions \cite{shoval2012evolutionarytradeoffspareto,pallasdies2021neuraloptimizationunderstanding}. Here, the empirical geometry suggests not a simple accuracy--cost compromise, but a structural floor imposed after modularity has already been constrained.

The combination of these findings has a clean interpretation: macroscopic modularity is selected as a structural invariant, not as a freely tunable parameter. The functional role of evolutionary refinement in the cost--modularity plane is then to find the cheapest networks consistent with that invariant. This pattern---fix the architectural template, minimize the resource cost within it---parallels the logic of biological wiring economy in neural systems, where wiring length, material cost, conduction delay, and communication efficiency are balanced rather than optimized in isolation \cite{raj2011thewiringeconomy,wang2016theinfluenceof,budd2012communicationandwiring}. The reservoirs in our experiment, evolved purely for predictive function on a chaotic task, recapitulate this principle without it being imposed as an objective.

This analogy is especially close to the wiring-economy view of cortical organization. Chklovskii, Schikorski, and Stevens argued that neural circuits are likely shaped by pressure to reduce wiring length, conduction delay, and material cost while preserving computational function \cite{Chklovskii2002wiringOptim}. Related work showed that wiring optimization can relate neuronal structure to function in the \emph{C. elegans} nervous system \cite{Chen2006wiringoptimization}. More recent work has extended this logic to artificial networks, showing that wiring-cost constraints can improve sparse neural-network performance and promote task-specific structural modules \cite{Zhang2025wiringeconomy}. The present reservoirs show a parallel version of this principle: predictive selection alone drives connection cost downward, but only within the modularity regime compatible with accurate forecasting.

\subsection{The reservoir as a model substrate for adaptive recurrent computation}

The standard reservoir-computing framework treats the recurrent network as a fixed random substrate and adapts only the readout \cite{jaeger2001echo,JaegerHaas2004esn,maass2002realtime,lukosevicius2009reservoircomputing,zhang2023asurveyon}. This is computationally useful, but it leaves aside a deeper question: how should a recurrent dynamical substrate itself be organized if it is selected for predictive function? By placing reservoir architecture under evolutionary selection, the present study makes that question explicit. The answer is not merely that larger or denser reservoirs perform better. Instead, the substrate is shaped along a specific set of axes: low-eigenvalue spectral modes are refined, modularity is constrained to an intermediate regime, and connection cost is reduced within that regime.

This interpretation returns to the biological motivation of the work. Biological systems do not generally learn their environments by constructing detached symbolic models. Gene regulatory networks, biochemical pathways, neural circuits, and other living substrates are embedded dynamical systems whose internal states are continuously conditioned by environmental history \cite{freddolino2012beyondhomeostasisa,gabaldasagarra2018recurrencebasedinformationprocessing,dehghani2024cellularcomplexity,seoane2019evolutionaryaspectsof}. The motivating idea behind this study was that such systems may act, in a broad sense, as predictive engines: they transform past inputs into present internal states that make future responses more adaptive. Reservoir computing is not a literal model of any one of these biological systems, but it provides a controlled setting in which the relation between prediction, recurrent dynamics, and substrate organization can be measured.

This perspective is complementary to recent work showing that measured cortical geometry, wiring, and function can be used directly as inductive biases for recurrent neural networks \cite{shakibaDehghani2026harnessingcorticalgeometrywiring}. In that setting, biological structure is imposed at initialization or through spatial and communicability constraints, and the resulting recurrent networks learn more effectively while developing modular and small-world organization. The present study asks the inverse question: if recurrent reservoirs are not given measured cortical priors, but are instead selected only for prediction, what structural constraints emerge? The convergence of these two directions supports a common view: recurrent computation benefits from structured substrates, whether those structures are inherited from biological organization or discovered through task-driven selection.

The evolved reservoirs should therefore be interpreted as model substrates for adaptive recurrent computation, not as direct biological replicas. Their significance lies in the structural logic that emerges under selection. Prediction does not favor arbitrary complexity. It favors a constrained dynamical class: enough recurrent capacity to sustain useful temporal memory, enough modular organization to preserve differentiated internal roles, enough integration to coordinate those roles, and enough cost pruning to avoid unnecessary connectivity. This combination---conserved architectural template, directional refinement of low-eigenvalue recurrent modes, modularity locking, and cost minimization within structural constraints---is consistent with a broader view of biological computation as physical, multiscale, modular, feedback-rich, and constrained by the relation between structure, dynamics, and function \cite{dehghani2018multiscalecontrol,dehghani2024physicalcomputing,dehghani2024cellularcomplexity}.

This framing also connects reservoir computing to emerging work on cellular, microbial, and chemical forms of information processing. Bacterial regulatory and metabolic systems, intracellular ion and cytoskeletal dynamics, and self-organizing reaction networks have all been proposed as possible substrates for reservoir-like computation \cite{jones2007liquidstate,niraula2024modeling,baltussen2024chemical,bartolome2026bacterialreservoir}. The present results do not show that these systems implement the same architecture. They show something more abstract: when a recurrent dynamical substrate is selected for prediction, its structure can become organized around a small number of interpretable constraints. That is the level at which reservoir computing becomes useful as a bridge between machine learning, dynamical systems, and biological computation.

\section{Outlook and conclusion}

These results suggest that predictive selection acts on recurrent reservoirs by stabilizing a task-suitable dynamical class rather than by simply favoring larger, denser, or arbitrarily complex networks. In the Kuramoto--Sivashinsky setting, evolutionary optimization improved prediction while producing a coordinated structural signature: extension of the low-error forecast horizon, a diminishing-return size--efficiency frontier, conservation of an SBM-like spectral envelope, directional refinement of the low-eigenvalue spectrum, locking of modularity to an intermediate band, and pruning of connection cost within that band.

The specificity of these signatures is likely to depend on the dynamical system being predicted. Spatially extended chaotic systems such as Kuramoto--Sivashinsky may place different demands on reservoir size, modularity, and slow spectral modes than low-dimensional chaotic systems, oscillatory signals, excitable media, or multiscale biological time series. Testing this dependence across target systems will be important for determining which reservoir constraints are task-specific and which reflect more general principles of adaptive recurrent prediction.

More broadly, the reservoir can be viewed not merely as a fixed random substrate with a trained readout, but as an adaptive recurrent medium whose structure reflects the predictive demands placed on it. This provides a population-level route to mechanistic interpretability for recurrent neural systems. Mechanistic interpretability here means identifying reproducible structural and dynamical links between substrate organization and function. In the present study, those links take the form of refined slow recurrent modes, modularity locking, and targeted pruning of connection cost. The patterns observed here emerged from selection on prediction error alone, without explicitly optimizing spectra, modularity, or wiring cost. Evolutionary reservoir computing therefore offers a way to study biological and artificial recurrent systems not only by asking whether they can predict, but by asking what structural constraints prediction selects in the substrate that performs it.

\section{Methods}
\subsection{Kuramoto--Sivashinsky time-series generation}

We used the Kuramoto--Sivashinsky equation \cite{Kuramoto1978KS,Sivashinsky1977KS} as a testbed for spatiotemporal chaos. The one-dimensional Kuramoto--Sivashinsky equation is
\begin{equation}
\frac{\partial u}{\partial t}
+
u \frac{\partial u}{\partial x}
+
\frac{\partial^{2}u}{\partial x^{2}}
+
\frac{\partial^{4}u}{\partial x^{4}}
=0,
\label{eq:ks_equation}
\end{equation}
where $u(x,t)$ denotes the system state at spatial position $x$ and time $t$. The system was simulated on a periodic spatial domain of length $L$, discretized into $n_x$ grid points. Unless otherwise specified, reported simulations used $n_x=64$, $L=22$, and time step $\Delta t=0.25$. Other tested KS sets also gave similar results (not reported).

Spatial derivatives were computed using Fourier spectral methods, where differentiation corresponds to multiplication by powers of the wave number $k$. In Fourier space, the linear term was represented using the multiplier
\begin{equation}
\mathcal{L}_k = k^2 - k^4,
\end{equation}
with the sign convention chosen to match the numerical implementation of Eq.~\ref{eq:ks_equation}. The nonlinear term was evaluated in physical space and transformed back to Fourier space at each time step.

Time integration was performed using the fourth-order exponential time-differencing Runge--Kutta method (ETDRK4), which treats the stiff linear term through the matrix exponential while integrating the nonlinear term explicitly \cite{Cox2002stiffode,Kassam2005edtrk}. The resulting simulated trajectories provided the target chaotic time series for reservoir training and prediction.

\subsection{Reservoir computing model}

Reservoir computing was used to predict the temporal evolution of the Kuramoto--Sivashinsky system. Each reservoir consisted of a recurrent network of $n_r$ nodes with adjacency matrix
\begin{equation}
\mathbf{A} \in \mathbb{R}^{n_r \times n_r}.
\end{equation}
The recurrent matrix was initialized as a sparse nonnegative weighted directed network. The desired average degree $d$ determined the sparsity level,
\begin{equation}
p = \frac{d}{n_r},
\label{eq:sparsity}
\end{equation}
where $p$ is the probability of a nonzero recurrent connection. The initial recurrent matrix $\mathbf{A}_0$ was generated as a sparse random matrix with nonzero weights sampled uniformly on $(0,1)$. The recurrent weights were then rescaled to impose the selected target spectral radius. Specifically, if
\begin{equation}
\lambda_{\max}
=
\max_\ell |\lambda_\ell(\mathbf{A}_0)|,
\end{equation}
and $\rho_{\mathrm{target}}$ denotes the spectral-radius hyperparameter selected by the genetic algorithm, then the recurrent matrix used in the reservoir dynamics was
\begin{equation}
\mathbf{A}
=
\rho_{\mathrm{target}}
\frac{\mathbf{A}_0}{\lambda_{\max}}.
\end{equation}
After this rescaling, the spectral radius of $\mathbf{A}$ is $\rho(\mathbf{A})=\rho_{\mathrm{target}}$.

The input weight matrix
\begin{equation}
\mathbf{W}_{\mathrm{in}} \in \mathbb{R}^{n_r \times n_u}
\end{equation}
mapped the $n_u$-dimensional input signal into the reservoir. The reservoir size was chosen as an integer multiple of $n_u$, so that each input dimension drove an equal block of $q=n_r/n_u$ reservoir nodes. Within each block, input weights were sampled independently from the uniform distribution $[-\sigma,\sigma]$, while entries outside the assigned block were set to zero.





At time $t$, the reservoir state $\mathbf{x}(t)\in \mathbb{R}^{n_r}$ evolved according to
\begin{equation}
\mathbf{x}(t)
=
f\!\left(
\mathbf{A}\mathbf{x}(t-1)
+
\mathbf{W}_{\mathrm{in}}\mathbf{u}(t)
\right),
\label{eq:reservoir_update}
\end{equation}
where $\mathbf{u}(t)$ is the input vector and $f(\cdot)$ is the elementwise hyperbolic tangent nonlinearity. The reservoir state was initialized at $\mathbf{x}(0)=\mathbf{0}$.

The reservoir output was computed using a linear readout,
\begin{equation}
\mathbf{y}(t)
=
\mathbf{W}_{\mathrm{out}}\mathbf{x}_{\mathrm{aug}}(t),
\label{eq:readout}
\end{equation}
where $\mathbf{x}_{\mathrm{aug}}(t)$ denotes the reservoir state, optionally augmented with nonlinear transformations or a bias term. The output matrix $\mathbf{W}_{\mathrm{out}}$ was the only component trained directly.

\subsection{Readout training}

The readout weights were trained using ridge regression.  Let
\begin{equation}
\mathbf{X}_{\mathrm{aug}}
=
\begin{bmatrix}
\mathbf{x}_{\mathrm{aug}}(1) & \mathbf{x}_{\mathrm{aug}}(2) & \cdots & \mathbf{x}_{\mathrm{aug}}(T)
\end{bmatrix}
\end{equation}
denote the matrix of augmented reservoir states collected during training, and let
\begin{equation}
\mathbf{Y}
=
\begin{bmatrix}
\mathbf{y}_{\mathrm{target}}(1) & \mathbf{y}_{\mathrm{target}}(2) & \cdots & \mathbf{y}_{\mathrm{target}}(T)
\end{bmatrix}
\end{equation}
denote the corresponding target outputs. The ridge-regression solution was
\begin{equation}
\mathbf{W}_{\mathrm{out}}
=
\mathbf{Y}
\mathbf{X}_{\mathrm{aug}}^{\top}
\left(
\mathbf{X}_{\mathrm{aug}}\mathbf{X}_{\mathrm{aug}}^{\top}
+
\beta \mathbf{I}
\right)^{-1},
\label{eq:ridge_regression}
\end{equation}
where $\beta$ is the regularization parameter.

During autonomous prediction, the reservoir was driven by its own previous output rather than by the true input. The prediction update was therefore:
\begin{align}
\mathbf{x}(t)
&=
f\!\left(
\mathbf{A}\mathbf{x}(t-1)
+
\mathbf{W}_{\mathrm{in}}\mathbf{y}(t-1)
\right),
\label{eq:prediction_update}
\\
\mathbf{y}(t)
&=
\mathbf{W}_{\mathrm{out}}\mathbf{x}_{\mathrm{aug}}(t).
\label{eq:prediction_output}
\end{align}
This autonomous mode was used to evaluate the ability of the reservoir to continue the chaotic trajectory beyond the training interval.

\subsection{Genetic algorithm optimization}
A genetic algorithm was used to optimize reservoir hyperparameters and structural features. Each individual in the population encoded a candidate reservoir configuration. The optimized parameters included reservoir size \(n_r\), spectral radius \(\rho(\mathbf{A})\), input scaling \(\sigma\), ridge regularization \(\beta\), and average degree \(d\) (or sparsity as defined by Eq.~\ref{eq:sparsity}).

The genetic algorithm proceeded through the following stages. First, an initial population of reservoirs was generated by sampling the encoded parameters from predefined ranges. Second, each reservoir was constructed, trained, and evaluated on the Kuramoto--Sivashinsky prediction task. Third, individuals were selected according to their fitness values. Fourth, crossover and mutation operators were applied to generate a new population of candidate reservoirs. This process was repeated across generations.

The optimization objective was to minimize a composite error score \(J\), defined below. Lower values of \(J\) corresponded to better predictive performance. Because reservoir construction and initialization involve stochasticity, all random seeds were controlled where appropriate to ensure reproducibility of the evolutionary runs.

\subsection{Prediction-error metrics}

Prediction accuracy was quantified using normalized error measures computed over a prediction horizon of length $T$. For output dimension $k$, the normalized root mean square error was
\begin{equation}
\mathrm{NRMSE}_k
=
\frac{
\sqrt{
\frac{1}{T}
\sum_{t=1}^{T}
\left(
y_k(t)-y_{\mathrm{target},k}(t)
\right)^2
}
}{
\sigma_{y_{\mathrm{target},k}}}.
\label{eq:nrmse}
\end{equation}

The normalized mean absolute error followed the implementation used in the evolutionary optimization. Both the predicted and target outputs were first min--max normalized using the minimum and maximum of the predicted output, and the normalized mean absolute deviation was computed as
\begin{equation}
\mathrm{NMAE}
=
\frac{1}{KT}
\sum_{k=1}^{K}
\sum_{t=1}^{T}
\left|
\tilde{y}_k(t)
-
\tilde{y}_{\mathrm{target},k}(t)
\right|,
\label{eq:nmae}
\end{equation}
where $\tilde{y}$ denotes the min--max normalized signals, \(K\) is the number of output dimensions and \(T\) is the number of prediction time points.

The composite fitness score was
\begin{equation}
J
=
\frac{
\mathrm{NMAE}
}{
\sum_{k=1}^{K}
\mathbf{1}\!\left[
\mathrm{NRMSE}_k < \varepsilon
\right]
},
\label{eq:fitness_function}
\end{equation}
where $\varepsilon$ is the NRMSE threshold and $\mathbf{1}[\cdot]$ is the indicator function. If no output dimension satisfied $\mathrm{NRMSE}_k<\varepsilon$, the denominator was zero and the individual was assigned $J=\infty$, corresponding to the lowest possible fitness under minimization. Individuals with lower finite \(J\) values were therefore considered more fit. This formulation reflects the fact that the Kuramoto--Sivashinsky system is spatially extended: a reservoir that fails to achieve acceptable accuracy on any output dimension is not capturing the underlying dynamics across the spatial field.

\subsection{Reservoir size--efficiency trade-off}
To quantify how evolutionary optimization shaped the relationship between reservoir size and predictive efficiency, we analyzed the empirical trade-off between reservoir size and prediction error across the evolved generations.

Each evaluated reservoir was represented as a point in a two-dimensional size--error plane,
\[
x_i = n_{r,i}, \qquad y_i = \log J_i ,
\]
where \(n_{r,i}\) is the reservoir size and \(J_i\) is the composite prediction error of the \(i\)th reservoir. The logarithm of \(J_i\) was used to emphasize relative differences among high-performing reservoirs and to make the lower-error region more visually interpretable. All evaluated reservoirs across all generations were included in this analysis, providing a comprehensive view of the size–accuracy landscape explored by evolution.

Efficiency was assessed using a Pareto‑front analysis in which both reservoir size and prediction error were treated as objectives to be minimized. A reservoir configuration \((x_i,y_i)\) was considered Pareto efficient if no other configuration \((x_j,y_j)\) satisfied
\[
x_j \leq x_i, \qquad y_j \leq y_i,
\]
with at least one strict inequality. Thus, the Pareto frontier identified reservoirs for which no alternative achieved both smaller size and lower prediction error.

This analysis provides a direct way to characterize the size--accuracy landscape explored by evolution. Points on the frontier correspond to nondominated compromises: smaller reservoirs can be efficient when they provide compact but less accurate solutions, whereas larger reservoirs remain efficient when they achieve lower errors that are not matched by smaller networks. The frontier therefore does not define a single optimal reservoir size, but rather the empirical boundary of attainable prediction error as a function of reservoir size.

To summarize the shape of this boundary, we fit an exponential curve,
\begin{equation}
f(x) = a e^{-b x} + c ,
\label{eq:expfit}
\end{equation}
to the Pareto-efficient points using nonlinear least squares. This fit provided a compact summary of the empirical size–accuracy frontier and allowed us to assess whether increasing reservoir size systematically lowered attainable error, and whether these improvements exhibited diminishing returns at larger  \(n_r\).

\subsection{Reservoir network representation}
Each reservoir was represented by its recurrent matrix \(\mathbf{A}\) after construction and spectral-radius scaling. Network analyses were performed directly on this saved weighted recurrent matrix, using the same matrix convention as in the reservoir update equation.


Because the reservoirs were directed and weighted, measures such as spectral analysis (random-walk Laplacian eigenvalues), connection density, path length, and connection cost were computed using directed weighted formulations. Community detection and modularity, whose implementations do not use edge direction, operated directly on the weighted adjacency matrix without manual symmetrization. All preprocessing steps were applied consistently across generations.

\subsection{Normalized random-walk Laplacian spectrum}
\label{sec:methods-laplacian}

To quantify reservoir structure, we computed the spectrum of the random-walk normalized graph Laplacian for each selected reservoir. For the directed weighted recurrent matrix \(\mathbf{A}\), the row weight of node \(i\) was defined as
\begin{equation}
d_i = \sum_j A_{ij}.
\end{equation}
These degrees formed the diagonal matrix
\begin{equation}
\mathbf{D} = \mathrm{diag}(d_1,d_2,\ldots,d_{n_r}).
\end{equation}
The random-walk normalized Laplacian was then defined as
\begin{equation}
\mathbf{L}_{\mathrm{rw}}
=
\mathbf{I}
-
\mathbf{D}^{-1}\mathbf{A}.
\label{eq:random_walk_laplacian}
\end{equation}

This random-walk normalization is appropriate for the present reservoirs because the reservoir connectivity matrices are directed and generally asymmetric. Unlike the symmetric normalized Laplacian, which is naturally defined for undirected graphs or symmetrized adjacency matrices, \(\mathbf{L}_{\mathrm{rw}}\) preserves the directed matrix structure by normalizing each row by its total row weight. In this way, the spectrum summarizes the relative transition structure induced by the directed recurrent substrate rather than the absolute magnitude of raw edge weights. For nonnegative directed graphs with positive row sum at each node, \(\mathbf{D}^{-1}\mathbf{A}\) is row-stochastic, so the eigenvalues of \(\mathbf{L}_{\mathrm{rw}}=\mathbf{I}-\mathbf{D}^{-1}\mathbf{A}\) lie in the disk centered at \(1\), providing a bounded spectral domain for comparison across networks with different sizes, densities, and degree heterogeneity \cite{chung1997spectral,Chung2005directedgraph}. Normalized Laplacian spectra have also been used as systems-level descriptors of neural network architecture because they capture global graph organization beyond node-wise summary statistics \cite{delange2014neuralspectrum}.

Eigenvalues of \(\mathbf{L}_{\mathrm{rw}}\) were computed using sparse eigensolvers from the ARPACK library \cite{lehoucq1998arpack}. Because directed weighted networks may produce complex eigenvalues, analyses used the real components of the eigenvalues as a summary of the reservoir's effective spectral organization. Reservoirs with unstable or ill-conditioned spectral decompositions were excluded from spectral population analyses according to the same criteria across generations.

The resulting eigenvalue set
\begin{equation}
\Lambda
=
\{\lambda_1,\lambda_2,\ldots,\lambda_m\}
\end{equation}
served as a spectral signature of the reservoir network. These spectra were used to compare reservoir structure across generations.

\subsection{Smoothed eigenvalue distributions}

To visualize the distribution of Laplacian eigenvalues, discrete spectra were converted into smoothed density functions. For eigenvalues \(\{\lambda_i\}_{i=1}^{m}\), the smoothed spectral density was computed as
\begin{equation}
\Gamma(x)
=
\frac{1}{m}
\sum_{i=1}^{m}
\frac{1}{\sqrt{2\pi s^2}}
\exp
\left(
-\frac{(x-\lambda_i)^2}{2s^2}
\right),
\label{eq:spectral_density}
\end{equation}
where \(s\) is the Gaussian smoothing bandwidth. The density was normalized so that the area under the curve summed to one. This procedure allowed visual comparison of eigenvalue distributions across networks and generations, following prior uses of smoothed normalized-Laplacian spectral plots for comparing network architectures \cite{delange2014neuralspectrum}.

\subsection{Network selection for spectral analysis}
To analyze the evolutionary progression of network structure while maintaining computational feasibility, we used a stratified sampling strategy across generations. We selected 10 generations evenly spaced across evolutionary time, including the initial population and the final generation containing at least 299 individuals in each generation.

Within each selected generation, reservoirs were ranked according to prediction error \(J\), with lower values indicating better performance. The population was divided into quartiles based on this ranking, and up to 20 reservoirs were sampled from each quartile (or all available reservoirs when fewer than 20 were present). This procedure yielded up to 80 reservoirs per generation. This stratification strategy ensured that the analysis captured not only elite performers but also the broader distribution of reservoir structures maintained within each generation.

\subsection{Principal component analysis of Laplacian spectra}

To compare Laplacian spectra across reservoirs of different sizes, each eigenvalue vector was sorted and interpolated to a common length \(q\). This produced a fixed-length spectral representation
\begin{equation}
\tilde{\Lambda}_i \in \mathbb{R}^{q}
\end{equation}
for each reservoir \(i\). The interpolated spectra were assembled into a matrix
\begin{equation}
\mathbf{S}
=
\begin{bmatrix}
\tilde{\Lambda}_1^{\top}
\\
\tilde{\Lambda}_2^{\top}
\\
\vdots
\\
\tilde{\Lambda}_M^{\top}
\end{bmatrix},
\end{equation}
where \(M\) is the number of sampled reservoirs.

Principal component analysis was then applied to \(\mathbf{S}\). The first two or three principal components were used to visualize the distribution of reservoirs in spectral space. Points were colored according to generation, allowing us to assess whether the reservoir population followed a coherent structural trajectory across evolutionary time.

\subsection{Spectral distance using optimal transport}

Pairwise distances between reservoir spectra were quantified using an optimal-transport distance  \cite{rubner2000earth,peyre2019computational}, reported throughout as Earth Mover's Distance (EMD). Each reservoir spectrum was treated as an empirical probability measure with equal mass assigned to each eigenvalue. Given two spectra
\begin{equation}
\Lambda^{(1)}
=
\{\lambda_1^{(1)},\lambda_2^{(1)},\ldots,\lambda_m^{(1)}\}
\end{equation}
and
\begin{equation}
\Lambda^{(2)}
=
\{\lambda_1^{(2)},\lambda_2^{(2)},\ldots,\lambda_n^{(2)}\},
\end{equation}
we assigned source and target weights
\begin{equation}
a_i = \frac{1}{m},
\qquad
b_j = \frac{1}{n},
\end{equation}
and computed the optimal transport cost
\begin{equation}
\mathrm{EMD}
\left(
\Lambda^{(1)},\Lambda^{(2)}
\right)
=
\min_{\gamma \in \Pi(a,b)}
\sum_{i=1}^{m}
\sum_{j=1}^{n}
\gamma_{ij}
C_{ij},
\label{eq:emd}
\end{equation}
where \(\gamma_{ij}\) is the transport plan, \(\Pi(a,b)\) is the set of couplings with marginals \(a\) and \(b\), and the cost matrix was defined using squared Euclidean distance,
\begin{equation}
C_{ij}
=
\left(
\lambda_i^{(1)}-\lambda_j^{(2)}
\right)^2 .
\end{equation}
Thus, the reported EMD values correspond to the unregularized optimal-transport cost between normalized empirical spectral measures with quadratic ground cost. This is equivalent to a squared 2-Wasserstein-type cost for the one-dimensional empirical distributions. The optimal transport problem was solved in Julia using the \texttt{emd}/\texttt{emd2} routines from \texttt{OptimalTransport.jl}/\texttt{ExactOptimalTransport.jl}, with \texttt{Tulip.jl} used as the linear-programming solver \cite{rubner2000earth,peyre2019computational,optimaltransportjl,Tanneau2021Tulip}. The resulting pairwise distance matrix was used to compare spectral similarity across reservoirs and generations.

\subsection{Community detection and modularity}

Community structure was estimated from the reservoir adjacency matrix using label propagation \cite{raghavan2007near}. For each reservoir, the adjacency matrix was converted to a weighted directed graph and community labels were assigned using the label-propagation algorithm \cite{raghavan2007near}. The resulting partition vector \(c_i\) assigns each reservoir node \(i\) to a community.

Modularity was then computed using the directed weighted Newman modularity \cite{newman2004finding,leicht2008community} implemented in \texttt{Graphs.jl}. For a directed graph, modularity is defined as
\begin{equation}
Q
=
\frac{1}{m}
\sum_{c}
\left[
e_c
-
\gamma
\frac{
K_c^{\mathrm{in}}
K_c^{\mathrm{out}}
}{m}
\right],
\label{eq:directed_modularity}
\end{equation}
where \(m\) is the total edge weight, \(e_c\) is the total weight of edges whose source and target nodes both lie within community \(c\), and \(K_c^{\mathrm{in}}\) and \(K_c^{\mathrm{out}}\) are the summed in- and out-degrees of nodes in community \(c\). We used \(\gamma=1\), corresponding to the traditional modularity definition. Modularity values were computed using \texttt{Graphs.modularity} with the reservoir weight matrix supplied through the \texttt{distmx} argument \cite{newman2004finding,leicht2008community,graphsjl}.

\subsection{Connection density, path length, and connection cost}

Connection density was computed as
\begin{equation}
\mathrm{density}
=
\frac{E}{n_r(n_r-1)},
\label{eq:density}
\end{equation}
where $E$ is the number of directed nonzero recurrent connections. For undirected symmetrized analyses, the corresponding undirected normalization was used.

Average path length was computed using shortest-path distances between all reachable ordered node pairs:
\begin{equation}
\ell
=
\frac{1}{|\mathcal{P}|}
\sum_{(i,j)\in \mathcal{P}}
d_{ij},
\label{eq:path_length}
\end{equation}
where $d_{ij}$ is the shortest directed path length from node $i$ to node $j$, and $\mathcal{P}$ is the set of node pairs connected by finite paths.

A regularized connection cost was defined as
\begin{equation}
C
=
\alpha
\sum_{ij}
|A_{ij}|
+
\beta_c
\,\mathrm{density}
+
\gamma
\,\ell,
\label{eq:connection_cost}
\end{equation}
where $\alpha$, $\beta_c$, and $\gamma$ are weighting parameters 
. This cost function combines total recurrent weight, connection density, and graph path length into a single measure of structural expense.

\subsection{Multi-objective analysis}
\label{sec:method-nsga}
To examine trade-offs among performance, modularity, and connection cost, each metric was normalized to the interval $[0,1]$. For a metric value $x$, the normalized value was
\begin{equation}
x_{\mathrm{norm}}
=
\frac{x - \min(x)}{\max(x) - \min(x)}.
\label{eq:minmax}
\end{equation}
Normalized metrics included prediction error, generation, modularity, and connection cost.

We constructed four objective functions to characterize different performance--structure trade-offs.

\paragraph{Objective 1: improvement across generations.}
Captures how prediction error decreases over evolutionary time by down-weighting later generations less strongly:
\begin{equation}
O_1
=
\frac{\mathrm{norm\_performance}}{1 + \mathrm{norm\_generation}}.
\label{eq:objective1}
\end{equation}

\paragraph{Objective 2: structural efficiency.}
Balances wiring cost against modular organization, favoring networks that achieve low cost with high modularity:
\begin{equation}
O_2
=
\frac{\mathrm{norm\_connection\_cost}}{1 + \mathrm{norm\_modularity}}.
\label{eq:objective2}
\end{equation}

\paragraph{Objective 3: performance--modularity relationship.}
Examines how prediction error varies with modular structure:
\begin{equation}
O_3
=
\frac{\mathrm{norm\_performance}}{1 + \mathrm{norm\_modularity}}.
\label{eq:objective3}
\end{equation}

\paragraph{Objective 4: performance--cost trade-off.}
Penalizes high prediction error more strongly when accompanied by high wiring cost:
\begin{equation}
O_4
=
\mathrm{norm\_performance}
\left(
1 + \mathrm{norm\_connection\_cost}
\right).
\label{eq:objective4}
\end{equation}

The additive $1+$ terms ensure that the objectives remain well-defined when normalized metrics take the value zero (avoiding division-by-zero in $O_1$--$O_3$) and that connection cost always contributes positively to $O_4$, rather than nullifying the objective when $\mathrm{norm\_connection\_cost} = 0$.

Multi-objective optimization was performed using the NSGA-II (Non-dominated Sorting Genetic Algorithm II) framework \cite{Deb2002NSGA} in Julia's \texttt{Metaheuristics.jl} package \cite{metaheuristics2022}. The algorithm was used to identify regions of the performance--structure landscape corresponding to favorable trade-offs among prediction error, modularity, and connection cost. Optimized positions in objective space were compared with the observed normalized data using Euclidean distance:
\begin{equation}
D
=
\sqrt{
\sum_i
\left(
x_{i,\mathrm{optimized}}
-
x_{i,\mathrm{observed}}
\right)^2
}.
\label{eq:euclidean_distance}
\end{equation}
The closest observed reservoirs were used to interpret the optimized trade-off solutions.

\emph{NSGA-II implementation:}
NSGA-II was configured with a population size of 1000, a crossover probability of 0.85, and a mutation probability of 0.5. All normalized metrics were bounded within $[0,1]$. Optimization proceeded for the full evolutionary run, and Pareto-optimal solutions were extracted for comparison with observed reservoirs.

For clarity, the NSGA-II procedure described here is independent of the genetic algorithm used to evolve the reservoir networks. The evolutionary algorithm generated the reservoir population, whereas NSGA-II was applied post hoc to the normalized metrics to characterize multi-objective trade-offs.

\subsection{Implementation and reproducibility}

Core reservoir simulations, training, prediction, and genetic algorithm optimization were implemented using numerical matrix operations and sparse linear algebra routines. Spectral graph analyses, community detection, optimal transport calculations, and multi-objective analyses were performed using \texttt{Julia}'s scientific computing libraries appropriate for sparse networks and graph analysis.

Random seeds were fixed for stochastic components of the simulations where possible, including reservoir initialization, genetic algorithm initialization, mutation, crossover, and network sampling. The same preprocessing and analysis pipeline was applied consistently across generations. Code and data required to reproduce the analyses will be made available in the associated repository.



\section{Code Availability}
The code used for simulation and network training, as well as the code for mechanistic probing of Reservoir architecture is available at \href{https://github.com/neurovium/AdaptiveReservoirComputing}{\texttt{Github: neurovium/AdaptiveReservoirComputing}}. All scripts required to reproduce the analyses and results presented in this study are provided, along with documentation and example usage.



\section{References}
\bibliography{EvoResComp}

\end{document}